\documentclass[runningheads]{style/llncs}
\usepackage{graphicx}

\usepackage{tikz}
\usepackage{style/comment}
\usepackage{amsmath,amssymb} 
\usepackage{color}

\usepackage[accsupp]{axessibility}  

\usepackage{caption}
\usepackage{subcaption}
\usepackage{booktabs}
\usepackage{multirow}
\usepackage{wrapfig,lipsum}
\usepackage{array}
\usepackage{float}
\usepackage{colortbl}
\usepackage{listings}
\usepackage{color}

\definecolor{dkgreen}{rgb}{0,0.6,0}
\definecolor{gray}{rgb}{0.5,0.5,0.5}
\definecolor{mauve}{rgb}{0.58,0,0.82}

\newcommand{\drule}{\specialrule{0.2pt}{1pt}{1pt}%
            \specialrule{0.2pt}{0pt}{\belowrulesep}%
            }
\newenvironment{rcases}
  {\left.\begin{aligned}}
  {\end{aligned}\right\rbrace}
  {\begin{list}{}%
          {\setlength{\leftmargin}{#1}}%
          \item[]%
  }
  {\end{list}}
  
\usepackage{kotex}

\definecolor{byzantine}{rgb}{0.74, 0.2, 0.64}

\usepackage[pagebackref=true,breaklinks=true,colorlinks,bookmarks=false]{hyperref}

\begin{document}
\pagestyle{headings}
\mainmatter
\def\ECCVSubNumber{1026}  

\title{Improving Test-Time Adaptation via Shift-agnostic Weight Regularization and Nearest Source Prototypes} 

\titlerunning{SWR \& NSP}
%
\author{Sungha Choi\thanks{Corresponding author.}\;\,
Seunghan Yang\;\,
Seokeon Choi\;\,
Sungrack Yun\\
}

\institute{Qualcomm AI Research\thanks{\vspace{-0.2cm}Qualcomm AI Research is an initiative of Qualcomm Technologies, Inc.}\\
\email{\{sunghac,seunghan,seokchoi,sungrack\}@qti.qualcomm.com}}

\authorrunning{S. Choi et al.}
%
\maketitle

\begin{abstract}
\vspace{-0.1cm}
This paper proposes a novel test-time adaptation strategy that adjusts the model pre-trained on the source domain using only unlabeled online data from the target domain to alleviate the performance degradation due to the distribution shift between the source and target domains. Adapting the entire model parameters using the unlabeled online data may be detrimental due to the erroneous signals from an unsupervised objective. To mitigate this problem, we propose a shift-agnostic weight regularization that encourages largely updating the model parameters sensitive to distribution shift while slightly updating those insensitive to the shift, during test-time adaptation. This regularization enables the model to quickly adapt to the target domain without performance degradation by utilizing the benefit of a high learning rate. In addition, we present an auxiliary task based on nearest source prototypes to align the source and target features, which helps reduce the distribution shift and leads to further performance improvement. We show that our method exhibits state-of-the-art performance on various standard benchmarks and even outperforms its supervised counterpart.

\vspace{-0.1cm}
\end{abstract}

\section{Introduction}
\vspace{-0.0cm}
After deep neural networks (DNNs) trained on a given dataset (\textit{i.e.,} source domain) are deployed to a new environment (\textit{i.e.,} target domain), the DNNs make predictions from the data in the target domain. However, in most cases, the distribution of the source and target domains varies significantly, which degrades the model's performance in the target domain. If the deployed model does not remain stationary during test time but adapts to the new environment using clues about unlabeled target data, its performance can be improved~\cite{sun2020test,wang2020tent,you2021test,mummadi2021test,iwasawa2021test,liang2020we,liu2021ttt++,zhang2022auxadapt}. 

\begin{figure*}[t!]
\centering
\includegraphics[width=\linewidth]{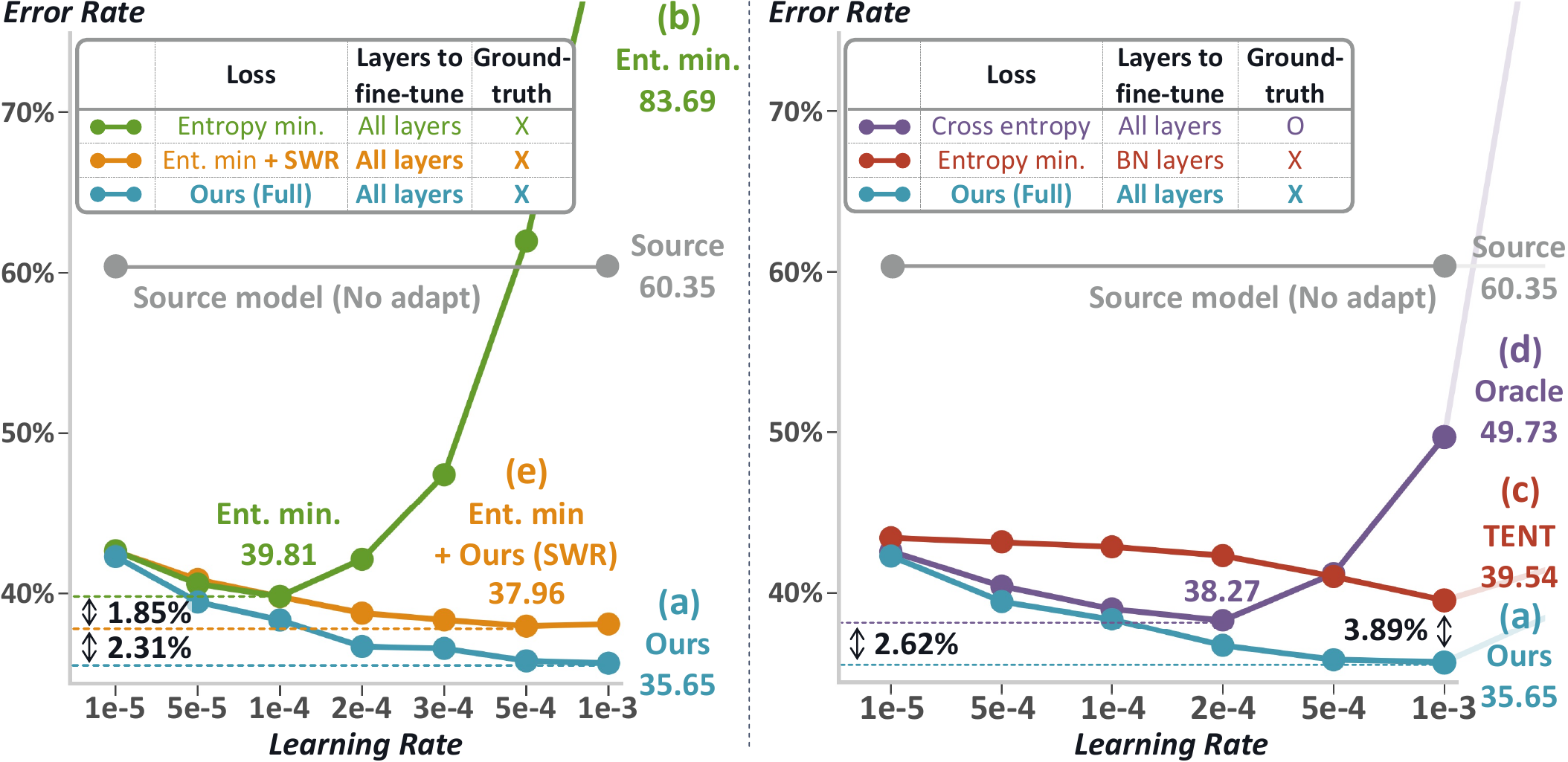}
\vspace{-0.55cm}
\caption{Comparison of average error ($\%$) between our approach and other methods with varying learning rates on CIFAR-100-C~\cite{hendrycks2018benchmarking}. The x- and y-axes are the learning rate and average error rate, respectively. (a) Our method significantly outperforms the other three methods: (b) updating the entire parameters with only entropy minimization, (c) the state-of-the-art method, TENT~\cite{wang2020tent}, and (d) a supervised method. (e) Our proposed SWR keeps the performance stable with the combining of entropy minimization even at higher learning rates: [1e-3, 1e-4].}
\label{fig:intro}
\vspace{-0.45cm}
\end{figure*}


Recently, several studies~\cite{wang2020tent,you2021test,mummadi2021test,iwasawa2021test} have proposed test-time adaptation to update the model during test time after model deployment. However, it is extremely challenging to adapt the model to the target domain with only unlabeled online data.
As shown in Fig.~\ref{fig:intro}(b), the adaptation of the entire model parameters may be detrimental due to the erroneous signals from the unsupervised objective such as entropy minimization~\cite{grandvalet2004semi,jain2018learning,long2016unsupervised,vu2019advent,wang2020tent}, and the performance may be highly dependent on the learning rate.
In addition, since the test-time adaptation can access unlabeled target data only once, and the adaptation proceeds simultaneously with the evaluation, updating all network parameters may result in overfitting~\cite{yosinski2014transferable,guo2019spottune}. 
Thus, several approaches present the methods to update only some part of the network architecture~\cite{wang2020tent,you2021test,mummadi2021test,iwasawa2021test} such as batch normalization~\cite{ioffe2015batch} or classifier layers. Especially, T3A~\cite{iwasawa2021test} proposes an optimization-free method to adapt only the classifier layers using unlabeled target data, and TENT~\cite{wang2020tent} updates batch statistics and affine parameters in the batch normalization layers by entropy minimization on unlabeled target data. 
However, updating only partial parameters or layers of the model may only result in marginal performance improvement, as shown in Fig.~\ref{fig:intro}(c).
Furthermore, such methods cannot be applied to the model architecture without a specific layer such as batch normalization or classifier layers.

Other approaches~\cite{sun2020test,liu2021ttt++} propose to jointly optimize a main task\footnote{\vspace{-0.1cm}refers to the ultimate objective of the model (\textit{e.g.,} classification).} and a self-supervised task, such as rotation prediction~\cite{gidaris2018unsupervised} or instance discrimination~\cite{he2020momentum,chen2020simple}, during pre-training in the source domain, and then update the model using only the self-supervised task during test time. In contrast to the unsupervised objective for the main task that highly depends on the model's prediction accuracy, the self-supervised task always obtains a proper supervisory signal. However, the self-supervised task may interfere with the main task if both tasks are not properly aligned~\cite{liu2021ttt++,su2020does,yu2020gradient}. In addition, these approaches cannot be applied to adapt arbitrary pre-trained models to the target domain since they require specific pre-training methods in the source domain.

To resolve these issues, we present two novel approaches for the test-time adaptation. First, we consider a shift-agnostic weight regularization (SWR) that enables the model to quickly adapt to the target domain, which is beneficial when updating the entire model parameters with a high learning rate. In contrast to Fig.~\ref{fig:intro}(b), the entropy minimization with the proposed SWR shows superior performance and less dependency on the learning rate choice, as shown in Fig.~\ref{fig:intro}(e). In terms of distribution shift, the SWR identifies the entire model parameters into shift-agnostic and shift-biased parameters, updating the former less and the latter more.
Second, we present an auxiliary task based on a non-parametric nearest source prototype (NSP) classifier, which pulls the target representation closer to its nearest source prototype. With the NSP classifier, both source and target representations can be well aligned, which significantly improves the performance of the main task. Our proposed method (Fig.~\ref{fig:intro}(a)) outperforms the state-of-the-art method~\cite{wang2020tent} (Fig.~\ref{fig:intro}(c)) and even the supervised method using ground-truth labels (Fig.~\ref{fig:intro}(d)).

Our method requires access to the source data to identify shift-agnostic and biased parameters and generate source prototypes before the model deployment, but it is applicable to any model regardless of its architecture or pre-training procedure. If a given model is pre-trained on open datasets, or if the source data owner deploys the model, source data is accessible before model deployment. In this case, our method significantly enhances the test-time adaptation capability by leveraging the source data without modifying the pre-trained model. Unlike TTT~\cite{sun2020test} and TTT\texttt{++}~\cite{liu2021ttt++}, we do not change the pre-training method of a given model, so our method can take benefit from any pre-trained strong models, such as AugMix~\cite{hendrycks2019augmix} (Table~\ref{tab_cifar}) or CORAL~\cite{sun2016deep} (Table~\ref{tab_domain}), as a good starting point for test-time adaptation. In these respects, we believe our method is practical. 

The major contributions of this paper can be summarized as follows
\vspace{-0.2cm}
\begin{list}{$\bullet$}{}  
    \item Two novel approaches for test-time adaptation are presented in this paper. The proposed SWR enables the model to quickly and reliably adapt to the target domain, and the NSP classifier aligns the source and target features to reduce the distribution shift, leading to further performance improvement.
\vspace{0.08cm}
    \item Our test-time adaptation method is model-agnostic and not dependent on the pre-training method in the source domain, and thus it can be applied to any pre-trained model. Therefore, our method can also complement other domain generalization approaches that mainly focus on the pre-training method in the source domain before model deployment.
\vspace{0.08cm}
    \item We show that our method achieves state-of-the-art performance through extensive experiments on CIFAR-10-C, CIFAR-100-C, ImageNet-C~\cite{hendrycks2018benchmarking} and domain generalization benchmarks including PACS~\cite{li2017deeper}, OfficeHome~\cite{venkateswara2017deep}, VLCS~\cite{Fang_2013_ICCV}, and TerraIncognita~\cite{beery2018recognition}. Especially, our method even outperforms its supervised counterpart on CIFAR-100-C dataset.
\end{list}
\section{Related Work}
\vspace{-0.0cm}
\subsection{Source-Free Domain Adaptation}
\vspace{-0.05cm}
Unsupervised domain adaptation (UDA)~\cite{sun2016deep,vu2019advent,tzeng2017adversarial,ganin2015unsupervised,saito2018maximum,hoffman2018cycada,ganin2016domain} assumes simultaneous access to both the source and target domains. Data is often distributed across multiple devices. In such cases, UDA requires data sharing for simultaneous access to all data. However, it is often impossible due to data privacy concerns, limited bandwidth, and computational cost. Source-free domain adaptation~\cite{liang2020we,Li_2020_CVPR,kundu2020universal,wang2021target,yang2021generalized,yeh2021sofa,Agarwal_2022_WACV} overcomes this challenge by adapting a source pre-trained model to the target domain using only unlabeled target data. These approaches focus on offline adaptation in which the same target sample is fed to the model multiple times during target adaptation, whereas our method concentrates on online adaptation.

\vspace{-0.3cm}
\subsection{Test-Time Adaptation and Training}
\vspace{-0.05cm}
Test-time adaptation focuses on online adaptation where all test data can be accessed only once, and adaptation is performed simultaneously with evaluation. 
More specifically, it forward propagates target samples through the model for evaluation and then backpropagates the error signal from the model's output in an unsupervised manner for training~\cite{wang2020tent}.
Several studies adopt self-supervised learning, such as rotation prediction~\cite{sun2020test} or instance discrimination~\cite{liu2021ttt++}, to jointly optimize the main and self-supervised tasks on the source domain and then optimize only the self-supervised task on the target domain.
However, these methods are not universally applicable to arbitrary pre-trained models as they require specific pre-training methods in the source domain. 
Recently, several methods~\cite{wang2020tent,iwasawa2021test,mummadi2021test,you2021test} have proposed model-agnostic test-time adaptation, which is independent of the pre-training method in the source domain.
TENT~\cite{wang2020tent} uses the batch statistics of the target domain and optimizes channel-wise affine parameters using entropy minimization loss. T3A~\cite{iwasawa2021test} proposes an optimization-free method that adjusts a pre-trained linear classifier by updating the prototype for each class during test time. However, since these methods update only partial parameters or layers of the model, such as the batch normalization~\cite{wang2020tent,mummadi2021test,you2021test} or classifier layer~\cite{iwasawa2021test}, they may be suboptimal for test-time adaptation.

\vspace{-0.3cm}
\subsection{Domain Generalization}
\vspace{-0.02cm}
Since UDA aims to adapt the model to the predefined target domain before model deployment, it is not suitable to guarantee generalization performance to other arbitrary target domains. On the other hand, domain generalization (DG) differs from UDA in that it assumes that the model accesses only the source domain during training time before model deployment and aims to improve the generalization capability in arbitrary unseen target domains. Numerous DG approaches using meta-learning~\cite{li2017learning,balaji2018metareg,li2019episodic}, normalization~\cite{pan2018two,seo2019learning,choi2021robustnet,Choi_2021_CVPR}, adversarial training~\cite{li2018domain,li2018deep,rahman2020correlation}, and data augmentation~\cite{zhou2020learning,volpi2018generalizing,gong2019dlow,li2021progressive} have been proposed to learn domain-agnostic feature representations for the target domain. However, these studies only focus on methods at training time before model deployment, whereas our method focuses on a test-time adaptation after model deployment.
\section{Proposed Method}
\vspace{-0.2cm}
\begin{figure}[!t]
\centering
\includegraphics[width=0.99\linewidth]{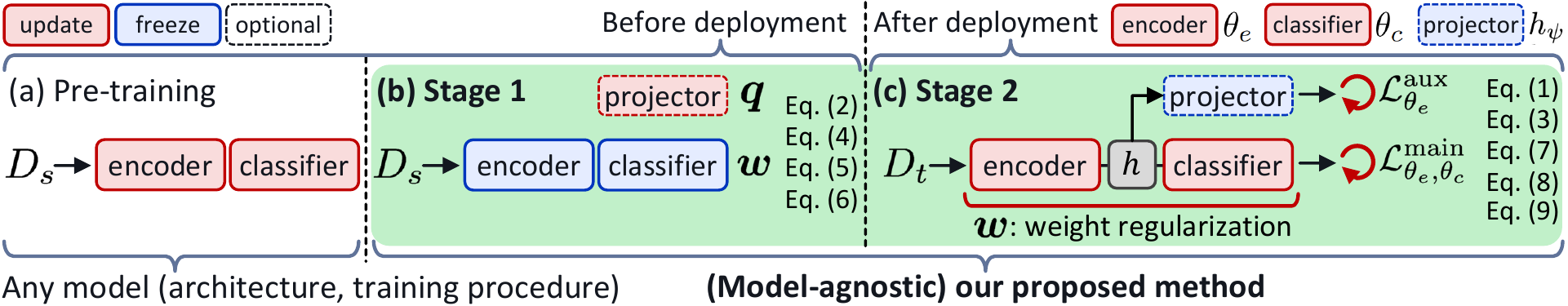}
\vspace{-0.1cm}
\caption{Our method consists of two stages: (b) and (c). (a) our method takes the pre-trained model in an off-the-shelf manner and (b) generates penalty vector $\boldsymbol{w}$ and source prototypes $\boldsymbol{q}$ while keeping the model \emph{frozen} before model deployment. After model deployment, (c) our method does not access labeled source data $D_s$ other than unlabeled online target data $D_t$ during test-time adaptation.}
\label{fig:method_overview}
\vspace{-0.5cm}
\end{figure}
Assume that the model parameters $\theta$ trained on the source domain consist of an encoder part $\theta_e$ and a classifier part $\theta_c$, as shown in Fig.~\ref{fig:method_overview}(c). 
After being deployed to the target domain, the model infers the class probability distribution of the target sample and then optimizes our proposed test-time adaptation loss $\mathcal{L}^{\text{target}}_{\theta_e,\theta_c}$. 
The overall loss of our proposed method is defined as
\vspace{-0.08cm}
\begin{equation} \label{eq_overall_loss}
\mathcal{L}^{\text{target}}_{\theta_e,\theta_c} =
\mathcal{L}^{\text{main}}_{\theta_e,\theta_c} +
\mathcal{L}^{\text{aux}}_{\theta_e} + 
\lambda_r\sum_l w_l\Vert\boldsymbol{\theta}_l-\boldsymbol{\theta}_l^*\Vert^2_2,
\vspace{-0.13cm}
\end{equation}
where $w_l$ denotes the $l$-th element of the penalty vector $\boldsymbol{w}$ used to control the update of the model parameters, $\boldsymbol{\theta}_l$ is the parameter vector of the $l$-th layer\footnote{\vspace{-0.05cm}denotes a part divided into torch.nn.Module units defined in Pytorch. The gradient vector of each layer can be easily obtained using torch.nn.module.parameters().} of the model, $\boldsymbol{\theta}^*$ is the parameters from the previous update step, 
$\lambda_r$ is the importance of the regularization term, 
and $\mathcal{L}^{\text{main}}_{\theta_e,\theta_c}$ and $\mathcal{L}^{\text{aux}}_{\theta_e}$ denote the main and auxiliary task losses, respectively. 
Optimizing the main task loss updates the entire model parameters $\theta_e$ and $\theta_c$, whereas optimizing the auxiliary task loss updates only the encoder part $\theta_e$. We first present a shift-agnostic weight regularization (SWR) 
and then describe an entropy objective of the main task. Finally, we propose an auxiliary task based on a nearest source prototype (NSP) classifier, which directly benefits the main task.

\vspace{-0.40cm}
\subsection{Shift-agnostic Weight Regularization}
\vspace{-0.09cm}
The main idea of the SWR is to impose different penalties for each parameter update during test-time adaptation, depending on the sensitivity of each model parameter to the distribution shift. Assuming that the distribution shift is mainly caused by color and blur shifts, we mimic the distribution shift using transformation techniques such as color distortion and Gaussian blur. Experiments on variations of the SWR, including the use of other transform functions, can be found in the supplementary Section B.

To obtain the penalty vector $\boldsymbol{w}$ specified in Eq.~(\ref{eq_overall_loss}), we first forward-propagate two input images (i.e., the original and its transformed image) through the source pre-trained model and then back-propagate the task loss (\textit{i.e.,} cross entropy) using the source labels to produce two sets $\boldsymbol{g}$ and $\boldsymbol{g}'$ of $L$ gradient vectors, respectively. Note that $L$ is the total number of layers in the model. Then the $l$-th element $w_l$ of the penalty vector $\boldsymbol{w}$ is calculated by employing the average cosine similarity $s_l$ between two gradient vectors, $\boldsymbol{g}_l$ and $\boldsymbol{g}'_l$ from $N$ source samples as

\vspace{-0.35cm}
\begin{equation} \label{eq_swr1}
\begin{split}
s_l&=\frac{1}{N}\sum_{i=1}^N\frac{\boldsymbol{g}^i_l\cdot{\boldsymbol{g}_{l}'^i}}{\Vert{\boldsymbol{g}^i_l}\Vert\Vert{{\boldsymbol{g}_{l}'^i}}\Vert}\in\mathbb{R}, \\
\boldsymbol{w}&=\left(\nu\left\lbrack s_1,\dots,s_l,\dots,s_{L} \right\rbrack\right)^2\in \mathbb{R}^L,
\vspace{-0.0cm}
\end{split}
\end{equation}
where $\nu\left\lbrack\cdot\right\rbrack$ denotes min-max normalization with the range of [0,1], 
$\boldsymbol{g}^i_l$ and $\boldsymbol{g}_{l}'^i$ denote the $l$-th gradient vectors for $i$-th source sample and its transformed sample, respectively. 
$N$ denotes the total number of samples. Note that the penalty vector $\boldsymbol{w}$ is obtained from a frozen pre-trained source model before model deployment. Therefore, this process is independent of the source model’s pre-training method and does not require source data after model deployment, as shown in Fig.~\ref{fig:method_overview}.

\begin{figure}[!t]
\centering
\includegraphics[width=\linewidth]{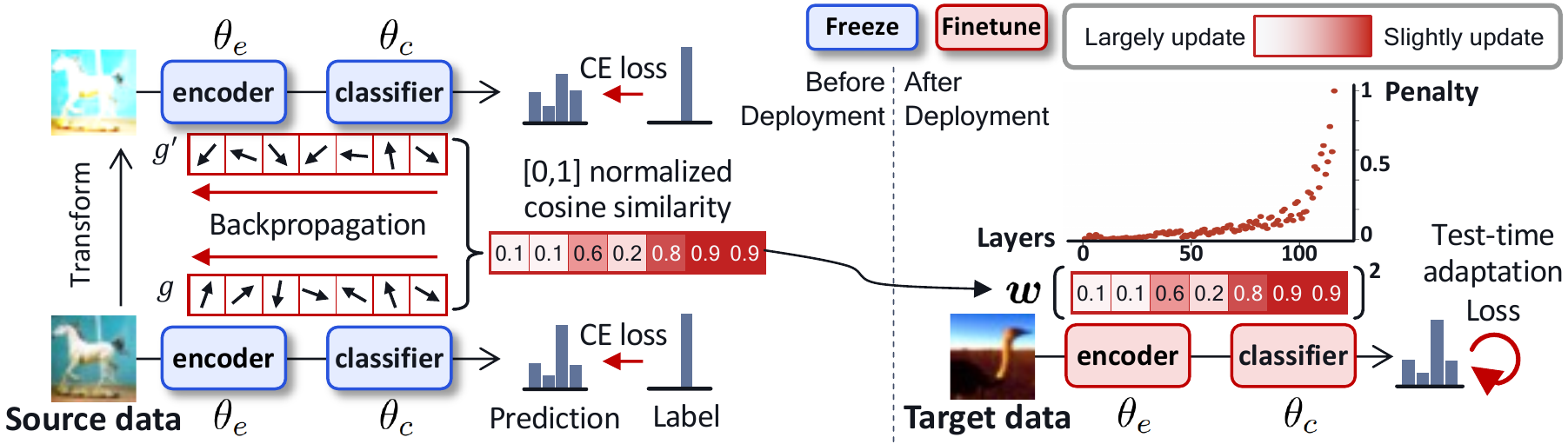}
\vspace{-0.4cm}
\caption{Overall process of our proposed SWR. We first obtain the penalty vector $\boldsymbol{w}$ before model deployment and then use it as layer-wise penalties to control the update of the model parameters at test-time adaptation after model deployment.}
\label{fig:method_swr}
\vspace{-0.45cm}
\end{figure}

As shown in Eq.~(\ref{eq_overall_loss}) and Fig.~\ref{fig:method_swr}, during test-time adaptation,
we apply the layer-wise penalty value $w_l$ to the difference between previous and current model parameters for each layer, 
and this controls the update of model parameters differently for each layer.
Therefore, the model parameters belonging to the layers with high cosine similarity between the two gradient vectors are considered shift-agnostic, and we less update them by imposing high penalties. 
Section~\ref{sec:exp_quick_adapt} experimentally shows that SWR takes advantage of using high learning rates to adapt the model to the target domain quickly.

\vspace{-0.38cm}
\subsection{Entropy Objective for the Main Task}
\vspace{-0.08cm}
The main task of the model $f_\theta$ is defined as the task performed by the parameters $\theta_e$ and $\theta_c$. The loss function for the main task during test time is built using the entropy of model predictions $\tilde{y}$ on test samples from the target distribution. 
We adopt information maximization loss~\cite{krause2010discriminative,shi2012information,hu2017learning}, validated in several test-time adaptation and source-free domain adaptation methods~\cite{liang2020we,wang2021target,mummadi2021test}, as an unsupervised learning objective for the main task. This loss consists of entropy minimization~\cite{wang2020tent,liang2020we,vu2019advent,springenberg2015unsupervised} and mean entropy maximization~\cite{NIPS2010_42998cf3,liang2020we,Assran_2021_ICCV,wang2021target} as
\vspace{-0.05cm}
\begin{equation} \label{eq_main_ent}
\vspace{-0.1cm}
\mathcal{L}^{\text{main}}_{\theta_e,\theta_c} = \lambda_{m_1}\frac{1}{N}\sum_{i=1}^N H(\tilde{y_i})-\lambda_{m_2} H(\bar{y}),
\vspace{-0.0cm}
\end{equation}
where $H(p)=-\sum_{k=1}^C p^k\log p^k$, $\bar{y}=\frac{1}{N}\sum_{i}\tilde{y_i}$, $\lambda_{m_1}$ and $\lambda_{m_2}$ indicate the importance of each term. The number of classes and the batch size are denoted by $C$ and $N$. Intuitively, entropy minimization makes individual predictions confident, and mean entropy maximization encourages average prediction within a batch to be close to the uniform distribution.


\vspace{-0.42cm}
\subsection{Auxiliary Task based on the Nearest Source Prototype}\label{sec_aux}
\vspace{-0.13cm}
Due to the distribution shift between the source and target domains, the target features deviate from the source features at test time.
To resolve this issue, we propose an auxiliary task based on the nearest source prototype (NSP) classifier, which pulls the target embeddings closer to their nearest source prototypes in the embedding space.
Eventually, optimizing the auxiliary task improves performance significantly since it directly supports the main task by aligning the source and target representations.
We first explain how to generate source prototypes and define the NSP classifier based on them.

\begin{figure}[!t]
\centering
\includegraphics[width=\linewidth]{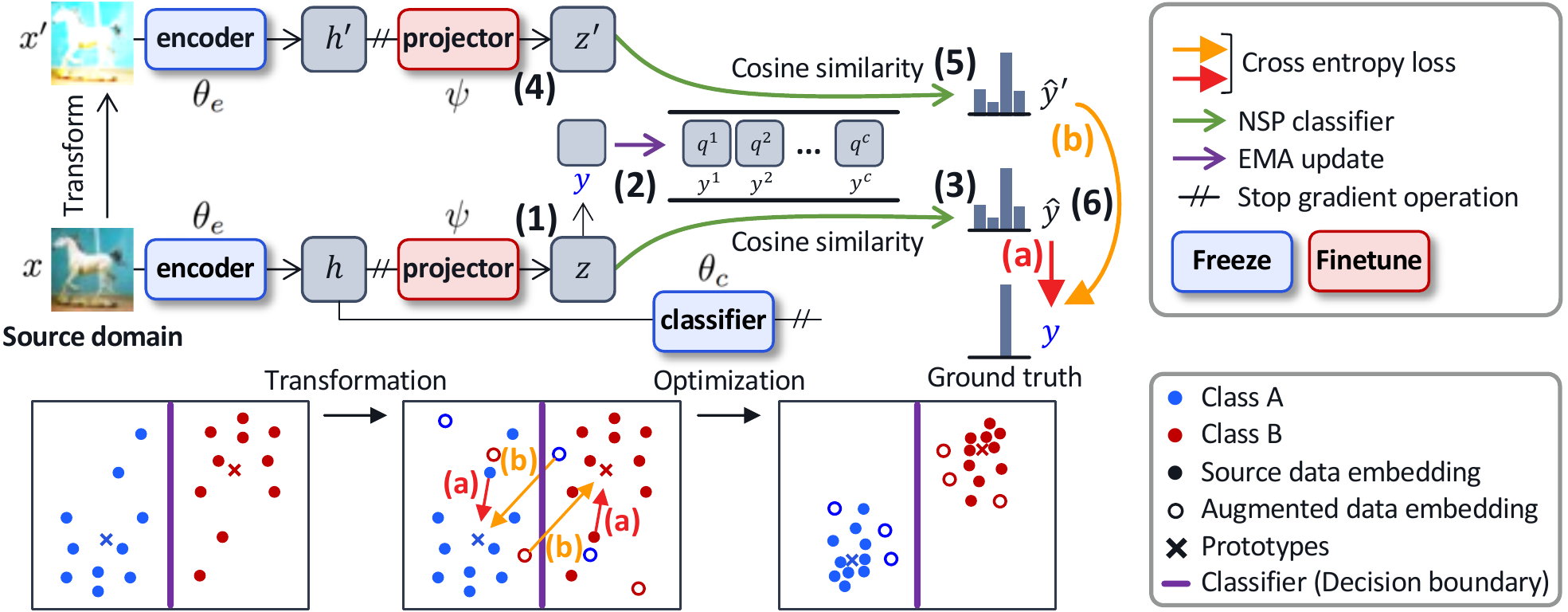}
\vspace{-0.55cm}
\caption{Source prototype generation phase before model deployment. First, we repeat steps (1) and (2) until prototypes of all classes are generated, then train the projector and update the source prototype at the same time through an iterative process from (1) to (6) on the source data. (a) and (b) pull the original source projection and its transformed source projection, respectively, such that they become closer to the nearest source prototype from the original one.}
\label{fig:method_pre}
\vspace{-0.45cm}
\end{figure}

\subsubsection{Source prototype generation}
\vspace{-0.5cm}
The source prototypes are defined as the averages over source embeddings for each class. 
As shown in Fig.~\ref{fig:method_pre}, we freeze the model $f_\theta$ trained on the source data and attach an additional projection layer $h_\psi$ behind the encoder $f_{\theta_e}$. The encoder $f_{\theta_e}$ infers the representation $\boldsymbol{h}$ from the source sample $x$, and the projector $h_\psi$ maps $\boldsymbol{h}$ to the projection $\boldsymbol{z}$ in another embedding space where the loss $\mathcal{L}^{\text{emb}}_{\psi}$ is applied as 
$\boldsymbol{z}=h_\psi(f_{\theta_e}(x))$.
The source prototype $\boldsymbol{q}^k_t$ for class $k$ is updated through exponential moving average (EMA) with the projection $\boldsymbol{z}^k_t$ of the source sample $(x, y^k)_{k\in[1,\text{C}]}$ at time $t$ during the optimization trajectory as
\vspace{-0.1cm}
\begin{equation} \label{eq_mean_entropy}
\boldsymbol{q}^k_t=\alpha\cdot \boldsymbol{q}^k_{t-1}+(1-\alpha)\cdot \boldsymbol{z}^k_t,
\vspace{-0.05cm}
\end{equation}
where $\alpha$=0.99 and $\boldsymbol{q}^k_0=\boldsymbol{z}^k_0$.

We define the NSP classifier as a non-parametric classifier. It measures the cosine similarity of a given target embedding to the source prototypes for all classes and then generates a class probability distribution $\hat{y}$ as
\vspace{-0.15cm}
\begin{equation} \label{eq_nsp_classifier}
\hat{y}=\sum_{k=1}^C\left(\frac{\text{exp}\left(S(\boldsymbol{z},\boldsymbol{q}^k)/\tau\right)}{\sum_{j=1}^C\text{exp}\left(S(\boldsymbol{z},\boldsymbol{q}^j)/\tau\right)}\right)y^k,
\vspace{-0.05cm}
\end{equation}
where $S(\cdot,\cdot)$ is a cosine similarity function, $S(\boldsymbol{a},\boldsymbol{b})=(\boldsymbol{a}\cdot\boldsymbol{b})/\Vert \boldsymbol{a}\Vert\Vert \boldsymbol{b}\Vert$, $\tau$ denotes a temperature that controls the sharpness of the distribution, and $y^k$ is the one-hot ground-truth label vector of $k$-th class. 
\begin{figure}[!t]
\centering
\includegraphics[width=\linewidth]{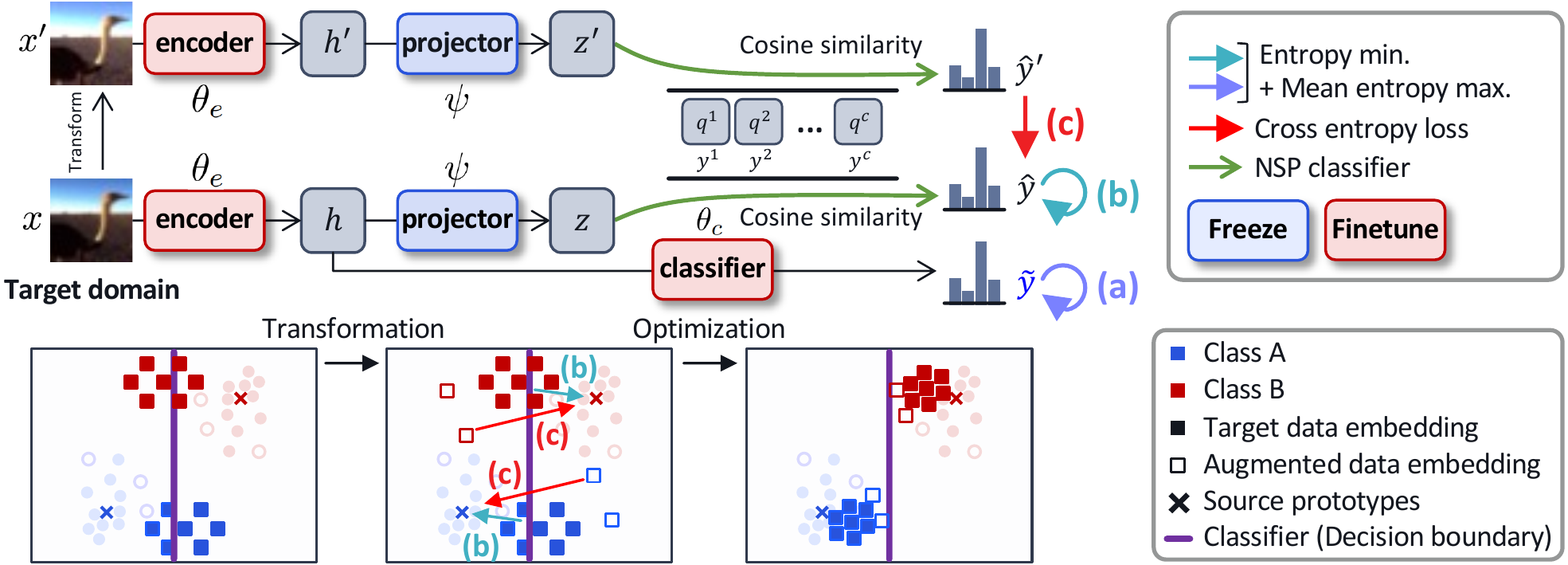}
\vspace{-0.55cm}
\caption{Test-time adaptation phase after model deployment. (a) main task loss. (b),(c) auxiliary task loss. (b) and (c) pull the original target projection and its transformed target projection, respectively, such that they become closer to the nearest source prototype from the original one.}
\label{fig:method_adapt}
\vspace{-0.5cm}
\end{figure}
In addition, inspired by recent self-supervised contrastive learning methods~\cite{chen2020simple,he2020momentum,bardes2021vicreg}, we enable the projector $h_\psi$ to learn transformation-invariant mapping. We obtain projection $\boldsymbol{z}'$ of the transformed source sample by 
$\boldsymbol{z}'=h_\psi(f_{\theta_e}(\mathcal{T}(x)))$, 
where $\mathcal{T}$($\cdot$) denotes an image transform function. The embedding loss $\mathcal{L}^{\text{emb}}_{\psi}$ consisting of two cross entropy loss terms is applied to the embedding space to train the projector $h_\psi$ as
\vspace{-0.25cm}
\begin{equation} \label{eq_whitening_loss}
\mathcal{L}^{\text{emb}}_{\psi}=\frac{1}{N}\sum_{i=1}^{N}\left(\text{CE}\left(y_i,{\hat{y}}_i\right)+\text{CE}\left(y_i,{\hat{y}'}_i\right)\right),
\vspace{-0.13cm}
\end{equation}
where $\text{CE}\left(p,q\right)=-\sum_{k=1}^C p^k\log q^k$, and $y_i$ is the ground-truth label of $i$-th source sample. Here, $\hat{y}$ and $\hat{y}'$ denote the outputs of the NSP classifier for the projections $\boldsymbol{z}$ and $\boldsymbol{z}'$ of the source sample and its transformed one, respectively. As shown in Fig.~\ref{fig:method_pre}, optimizing the embedding loss encourages the projector $h_\psi$ to learn a mapping that pulls the projections belonging to the same class closer together and pushes source prototypes farther away from each other.

Note that this process is applied to a frozen pre-trained source model and completed before model deployment. Therefore, it is model-agnostic and does not require source data during test time.

\vspace{-0.1cm}
\subsubsection{Auxiliary task loss at test time}
Once the source prototypes are generated and the projection layer is trained, we can deploy the model and then jointly optimize both main and auxiliary tasks on unlabeled online data. The auxiliary task loss $\mathcal{L}^{\text{aux}}_{\theta_e}$ consists of two objective functions: the entropy objective $\mathcal{L}^{\text{aux\_ent}}_{\theta_e}$ using the entropy of the NSP classifier's prediction $\hat{y}$, and the self-supervised loss $\mathcal{L}^{\text{aux\_sel}}_{\theta_e}$ that encourages the model's encoder $f_{\theta_e}$ to learn transformation-invariant mappings as

\vspace{-0.2cm}
\begin{equation} \label{eq_aux_loss}
\mathcal{L}^{\text{aux}}_{\theta_e} = \mathcal{L}^{\text{aux\_ent}}_{\theta_e} + \lambda_s\mathcal{L}^{\text{aux\_sel}}_{\theta_e},
\vspace{-0.0cm}
\end{equation}
where $\lambda_s$ denotes the importance of the self-supervised loss term. Similarly to Eq.~(\ref{eq_main_ent}), the entropy objective is built by using the entropy of the prediction $\hat{y}$ of the NSP classifier on the target sample as
\vspace{-0.1cm}
\begin{equation} \label{eq_aux_ent}
\mathcal{L}^{\text{aux\_ent}}_{\theta_e} = \lambda_{a_1}\frac{1}{N}\sum_{i=1}^{N}H(\hat{y_i})-\lambda_{a_2}H(\bar{y}),
\vspace{-0.05cm}
\end{equation}
where $N$ is batch size, $\lambda_{a_1}$ and $\lambda_{a_2}$ indicate the importance of each term, $H(p)=-\sum_{k=1}^C p^k\log p^k$, and $\bar{y}=\frac{1}{N}\sum_{i=1}^{N}\hat{y_i}$.
The self-supervised loss is applied to the prediction $\hat{y}'$ of the NSP classifier on the transformed target sample as
\vspace{-0.1cm}
\begin{equation} \label{eq_aux_sel}
\mathcal{L}^{\text{aux\_sel}}_{\theta_e}=-\frac{1}{N}\sum_{i=1}^{N}\sum_{k=1}^C\hat{y}_i^k\log {\hat{y}'^k_i}.
\vspace{-0.05cm}
\end{equation}
As shown in Fig.~\ref{fig:method_adapt}, the entropy objective function (Fig.~\ref{fig:method_adapt}(b)) pulls the projection $\boldsymbol{z}$ of the target sample to move closer to its nearest source prototype, and the self-supervised objective (Fig.~\ref{fig:method_adapt}(c)) encourages the projection $\boldsymbol{z}'$ of the transformed target sample to get closer to the same target as $\boldsymbol{z}$.

\vspace{-0.25cm}
\section{Experiments}
\vspace{-0.15cm}
This section describes the experimental setup, implementation details, and the 
experimental results of the comparisons with other state-of-the-art methods 
in test-time adaptation. We also show that generalization performance can 
be further improved  
by combining our proposed method with an existing domain generalization strategy that mainly focuses on training time in the source domain.

\vspace{-0.35cm}
\subsection{Experimental Setup}
\vspace{-0.05cm}
Following TENT~\cite{wang2020tent} and T3A~\cite{iwasawa2021test}, all experiments in this paper are conducted on the online adaptation setting, where adaptation is performed concurrently with evaluation at test time without seeing the same data twice or more.
After a prediction is obtained, the model is updated via back-propagation.
We evaluate our proposed method on \textbf{CIFAR-10-C}, \textbf{CIFAR-100-C}, \textbf{ImageNet-C}\footnote{\vspace{-0.45cm}Experiments on ImageNet-C are in the supplementary Section B.}~\cite{hendrycks2018benchmarking} and four domain generalization benchmarks such as \textbf{PACS}~\cite{li2017deeper}, \textbf{OfficeHome}~\cite{venkateswara2017deep}, \textbf{VLCS}~\cite{Fang_2013_ICCV}, and \textbf{TerraIncognita}~\cite{beery2018recognition}. 
Since our method can be used independently of the backbone networks and its pre-training method, 
we apply our method to publicly available pre-trained models for evaluation. We perform experiments on CIFAR datasets using WideResNet-28-10~\cite{zagoruyko2016wide} and WideResNet-40-2~\cite{zagoruyko2016wide} as backbone networks, based on RobustBench~\cite{croce2020robustbench}. In the domain generalization setup, we use ResNet-50~\cite{he2016deep} without the batch normalization layer, which is the default setting of DomainBed~\cite{gulrajani2020search}, DG benchmark framework.
CIFAR-10/100 dataset~\cite{krizhevsky2009learning} contains 50k images for training and 10k images for testing. Corruptions such as noise, blur, weather, and digital are applied to 10k images from CIFAR test set to create CIFAR-C test images. For test-time adaptation, 50k images for CIFAR training set are defined as the source domain, and 10k images for CIFAR-C test set are defined as the target domain.

\vspace{-0.4cm}
\subsection{Implementation details}
\vspace{-0.1cm}
We integrate our proposed method within the frameworks officially provided by other state-of-the-art methods~\cite{wang2020tent,liu2021ttt++,iwasawa2021test} for fair comparisons.
Specifically, different frameworks are used for each experiment as follows: \href{https://github.com/DequanWang/tent}{TENT framework}~\cite{wang2020tent} for all experiments with WRN-40-2 and WRN-28-10 backbone networks on CIFAR-10/100-C, \href{https://github.com/vita-epfl/ttt-plus-plus}{TTT\texttt{++} framework}~\cite{liu2021ttt++} for all experiments with ResNet-50 on CIFAR-10/100-C, and \href{https://github.com/matsuolab/T3A}{T3A framework}~\cite{iwasawa2021test} for all domain generalization benchmarks. For experiments on CIFAR, we follow the default values provided by each framework for experimental settings such as batch size and optimizer.

Color distortion, random grayscale and Gaussian blurring are used as the image transformations specified in Fig.~\ref{fig:method_swr} and Fig.~\ref{fig:method_adapt}, and random cropping and random horizontal flipping are additionally applied for the image transformations in Fig.~\ref{fig:method_pre}.
We use batch statistics on test data instead of using running estimates. The hyper-parameters are empirically set as $\lambda_{m_1}$=0.2, $\lambda_{a_1}$=0.8
$\lambda_{m_2}$=0.25, $\lambda_{a_2}$=0.25, $\lambda_s$=0.1, $\lambda_r$=250, and softmax temperature $\tau$=0.1. The epoch for training the projector is 20, and $N$=1024 in Eq.~(\ref{eq_swr1}). Since these hyper-parameters are not sensitive to the backbone and datasets, they are fixed without individual tuning in most experiments in this paper unless noted otherwise.
The projector as described in Section~\ref{sec_aux} can be configured as a single- or multi-layer perceptron (MLP). The MLP consists of a linear layer followed by batch normalization~\cite{ioffe2015batch}, ReLU~\cite{nair2010rectified}, and a final linear layer with output dimension 512. The performance change according to the projector configuration is shown in Table~\ref{tab_projection_layer}, and the detailed architecture is described in the supplementary Section C.


\begin{table}[t]
\caption{Comparison with other methods. * denotes the reported results from the original paper, and the others are reproduced values in our environment based on the official framework provided by TENT~\cite{wang2020tent} and TTT\texttt{++}~\cite{liu2021ttt++}. Source denotes the source pre-trained model without test-time adaptation.} 
\vspace{-0.2cm}
\begin{subtable}[h]{\textwidth}
    \centering
        \setlength\tabcolsep{2pt}
        \renewcommand{\arraystretch}{0.8}
        \footnotesize
        \caption{Comparison of error rate ($\%$) on CIFAR-100-C with severity level 5}
        \vspace{-0.1cm}
        \resizebox{\textwidth}{!}{
        \begin{tabular}{c|c|c|ccccccccccccccc}
        \toprule
        \rule{0pt}{5pt}\textbf{Backbone} & \textbf{Methods} & \textbf{Avg. err} & \textbf{Gaus.} & \textbf{Shot} & \textbf{Impu.} & \textbf{Defo.} & \textbf{Glas.} & \textbf{Moti.} & \textbf{Zoom} & \textbf{Snow} & \textbf{Fros.} & \, \textbf{Fog} \, & \textbf{Brig.} & \textbf{Cont.} & \textbf{Elas.} & \textbf{Pixe.} & \textbf{Jpeg} \\
        \drule
        \multirow{4}{*}{\shortstack{\\ \\ \\ \\ WRN-40-2\\(AugMix)\\~\cite{zagoruyko2016wide,hendrycks2019augmix}}} & Source & 46.75 & 65.7 & 60.1 & 59.1 & 32.0 & 51.0 & 33.6 & 32.3 & 41.4 & 45.2 & 51.4 & 31.6 & 55.5 & 40.3 & 59.7 & 42.4 \\
        \arrayrulecolor{black!50}\cmidrule{2-18}
        & TENT~\cite{wang2020tent} & 35.53 & 40.1 & 39.5 & 42.0 & 29.6 & 41.9 & 30.7 & 29.7 & 34.5 & 34.8 & 39.1 & 27.5 & 32.9 & 37.6 & 32.8 & 40.3  \\
        \cmidrule{2-18}
        & Core*~\cite{you2021test} & 35.30 & 39.8 & 39.3 & 41.5 & 29.5 & 41.7 & 30.6 & 29.8 & 34.2 & 34.9 & 38.6 & 27.5 & 32.6 & 37.1 & 32.7 & 40.1  \\
        \cmidrule{2-18}
        & \cellcolor[gray]{0.9}\textbf{Ours} & \textbf{32.71} & \textbf{37.6} & \textbf{36.6} & \textbf{35.1} & \textbf{28.0} & \textbf{39.5} & \textbf{28.7} & \textbf{28.5} & \textbf{31.3} & \textbf{32.6} & \textbf{34.4} & \textbf{26.3} & \textbf{29.0} & \textbf{35.5} & \textbf{30.1} & \textbf{37.7}  \\
        \arrayrulecolor{black!100}\midrule
        \multirow{6}{*}{\shortstack{\\ \\ \\ \\ \\ \\ \\ \\ \\ ResNet-50\\~\cite{he2016deep}}}
        & Source & 60.35 & 80.8 & 77.8 & 87.8 & 39.6 & 82.3 & 54.2 & 38.4 & 54.6 & 60.2 & 68.1 & 28.9 & 50.9 & 59.5 & 72.3 & 50.0 \\
        \arrayrulecolor{black!50}\cmidrule{2-18}
        & SHOT~\cite{liang2020we}
        & 43.53 & 49.0 & 47.1 & 61.2 & 33.8 & 58.3 & 41.0 & 31.3 & 45.0 & 42.0 & 52.0 & 29.7 & 33.4 & 47.9 & 41.8 & 39.4  \\
        \cmidrule{2-18}
        & TFA~\cite{liu2021ttt++} & 44.13 & 49.0 & 47.0 & 61.4 & 34.2 & 58.9 & 41.5 & 32.1 & 46.8 & 43.0 & 54.6 & 31.2 & 33.7 & 48.9 & 39.8 & 39.7  \\
        \cmidrule{2-18}
        & TTT\texttt{++}~\cite{liu2021ttt++} & 44.38 & 50.2 & 47.7 & 66.1 & 35.8 & 61.0 & 38.7 & 35.0 & 44.6 & 43.8 & 48.6 & 28.8 & 30.8 & 49.9 & 39.2 & 45.5  \\
        \cmidrule{2-18}
        & TENT~\cite{wang2020tent}
        & 39.54 & 43.8 & 42.0 & 54.1 & 31.2 & 51.7 & 37.0 & 29.9 & 42.3 & 39.6 & 45.6 & 30.1 & 30.9 & 44.5 & 34.2 & 36.4  \\
        \cmidrule{2-18}
        & \cellcolor[gray]{0.9}\textbf{Ours} & \textbf{35.65} & \textbf{40.0} & \textbf{38.4} & \textbf{46.3} & \textbf{29.3} & \textbf{46.0} & \textbf{32.5} & \textbf{27.9} & \textbf{37.3} & \textbf{36.6} & \textbf{37.3} & \textbf{27.5} & \textbf{28.8} & \textbf{41.0} & \textbf{31.4} & \textbf{34.7}  \\
        \arrayrulecolor{black!100}\bottomrule
        \end{tabular}}
    \label{tab_cifar100}
\end{subtable}
\vfill
\begin{subtable}[h]{\textwidth}
    \centering
        \setlength\tabcolsep{2pt}
        \renewcommand{\arraystretch}{0.8}
        \vspace{-0.2cm}
        \footnotesize
        \caption{Comparison of error rate ($\%$) on CIFAR-10-C with severity level 5}
        \vspace{-0.1cm}
        \resizebox{\textwidth}{!}{
        \begin{tabular}{c|c|c|ccccccccccccccc}
        \toprule
        \rule{0pt}{5pt}\textbf{Backbone} & \textbf{Methods} & \textbf{Avg. err} & \textbf{Gaus.} & \textbf{Shot} & \textbf{Impu.} & \textbf{Defo.} & \textbf{Glas.} & \textbf{Moti.} & \textbf{Zoom} & \textbf{Snow} & \textbf{Fros.} & \, \textbf{Fog} \, & \textbf{Brig.} & \textbf{Cont.} & \textbf{Elas.} & \textbf{Pixe.} & \textbf{Jpeg} \\
        \drule
        \multirow{3}{*}{\shortstack{\\WRN-40-2\\(AugMix)\\~\cite{zagoruyko2016wide,hendrycks2019augmix}}} & Source & 18.27 & 28.8 & 23.0 & 26.2 & 9.5 & 20.6 & 10.6 & 9.3 & 14.2 & 15.3 & 17.5 & 7.6 & 20.9 & 14.7 & 41.3 & 14.7 \\
        \arrayrulecolor{black!50}\cmidrule{2-18}
        & TENT~\cite{wang2020tent} & 12.08 & 15.6 & 13.2 & 18.8 & 7.9 & 18.2 & 9.0 & 8.0 & 10.4 & 10.9 & 12.4 & 6.7 & 10.0 & 14.0 & 11.4 & 14.8  \\
        \cmidrule{2-18}
        & \cellcolor[gray]{0.9}\textbf{Ours} & \textbf{10.37} & \textbf{13.1} & \textbf{11.4}& \textbf{14.7} & \textbf{7.4} & \textbf{15.8} & \textbf{8.3} & \textbf{7.4} & \textbf{9.2} & \textbf{9.4} & \textbf{9.5} & \textbf{6.3} & \textbf{7.8} & \textbf{13.1} & \textbf{9.4} & \textbf{12.9}  \\
        \arrayrulecolor{black!100}\midrule
        \multirow{4}{*}{\shortstack{\\ \\ \\ \\ \\WRN-28-10\\~\cite{zagoruyko2016wide}}}
        & Source & 43.51 & 72.3 & 65.7 & 72.9 & 49.9 & 54.3 & 34.8 & 42.0 & 25.1 & 41.3 & 26.0 & 9.3 & 46.7 & 26.6 & 58.5 & 30.3 \\
        \cmidrule{2-18}
        & TENT~\cite{wang2020tent} & 18.58 & 24.8 & 23.5 & 33.1 & 11.9 & 31.8 & 13.7 & 10.8 & 15.9 & 16.2 & 13.7 & 7.9 & 12.0 & 22.0 & 17.3 & 24.2  \\
        \arrayrulecolor{black!50}\cmidrule{2-18}
        & Core*~\cite{you2021test} & 16.80 & 22.5 & 20.3 & 29.8 & 11.0 & 29.2 & 12.3 & \textbf{10.2} & 14.4 & 14.8 & 12.4 & 7.7 & 10.6 & 20.4 & 15.3 & 21.4  \\
        \cmidrule{2-18}
        & \cellcolor[gray]{0.9}\textbf{Ours} & \textbf{15.70} & \textbf{20.1} & \textbf{18.4} & \textbf{26.2} & \textbf{10.8} & \textbf{28.9} & \textbf{12.1} & \textbf{10.2} & \textbf{13.7} & \textbf{13.9} & \textbf{11.1} & \textbf{7.6} & \textbf{8.8} & \textbf{20.2} & \textbf{14.2} & \textbf{19.4}  \\
        \arrayrulecolor{black!100}\midrule
        \multirow{6}{*}{\shortstack{\\ \\ \\ \\ \\ \\ \\ \\ \\ResNet-50\\~\cite{he2016deep}}}
        & Source & 29.14 & 48.7 & 44.0 & 57.0 & 11.8 & 50.8 & 23.4 & 10.8 & 21.9 & 28.2 & 29.4 & \textbf{7.0} & 13.13 & 23.4 & 47.9 & 19.5 \\
        \arrayrulecolor{black!50}\cmidrule{2-18}
        & SHOT~\cite{liang2020we}
        & 16.19 & 20.0 & 18.8 & 29.6 & 9.9 & 27.1 & 15.0 & 8.5 & 15.4 & 14.5 & 19.8 & 7.3 & 8.5 & 18.7 & 15.8 & 14.0  \\
        \cmidrule{2-18}
        & TFA~\cite{liu2021ttt++} & 15.97 & 18.8 & 17.9 & 29.2 & 9.8 & 27.3 & 14.6 & 8.0 & 16.0 & 14.0 & 20.3 & 7.8 & 8.6 & 19.4 & 14.1 & 13.9  \\
        \cmidrule{2-18}
        & TTT\texttt{++}~\cite{liu2021ttt++} & 15.82 & 18.0 & 17.1 & 30.8 & 10.4 & 29.9 & 13.0 & 9.9 & 14.8 & 14.1 & 15.8 & \textbf{7.0} & 7.8 & 19.3 & 12.7 & 16.4  \\
        \cmidrule{2-18}
        & TENT~\cite{wang2020tent}
        & 14.02 & 16.0 & 14.5 & 24.7 & 9.1 & 23.5 & 12.6 & 7.6 & 14.3 & 13.1 & 16.8 & 8.2 & 8.0 & 18.1 & 10.8 & 13.3  \\
        \cmidrule{2-18}
        & \cellcolor[gray]{0.9}\textbf{Ours} & \textbf{12.52} & \textbf{14.1} & \textbf{13.4} & \textbf{20.9} & \textbf{8.3} & \textbf{20.7} & \textbf{11.2} & \textbf{7.3} & \textbf{12.4} & \textbf{11.7} & \textbf{14.4} & 7.3 & \textbf{7.4} & \textbf{16.5} & \textbf{9.7} & \textbf{12.4}  \\
        \arrayrulecolor{black!100}\bottomrule
        \end{tabular}}
    \label{tab_cifar10}
\end{subtable}
\vfill
\begin{subtable}[h]{\textwidth}
    \centering
        \setlength\tabcolsep{6pt}
        \renewcommand{\arraystretch}{0.3}
        \vspace{-0.2cm}
        \footnotesize
        \caption{Comparison of average error ($\%$) on CIFAR-100-C  with all severity levels}
        \vspace{-0.1cm}
        \resizebox{310pt}{!}{
        \label{tab_all_level}
        \begin{tabular}{c|c|c|c|c|c|c|c|c|c|c|c}
        \toprule
        \textbf{Methods} & \textbf{Backbone} & \multicolumn{2}{c|}{\textbf{Lv.5}} & \multicolumn{2}{c|}{\textbf{Lv.4}} & \multicolumn{2}{c|}{\textbf{Lv.3}} & \multicolumn{2}{c|}{\textbf{Lv.2}} & \multicolumn{2}{c}{\textbf{Lv.1}} \\ 
        \arrayrulecolor{black!100}\midrule
        TENT~\cite{wang2020tent} & \multirow{2}{*}{\shortstack{\\ResNet-50}} & 39.54 & \multirow{2}{*}{\shortstack{\\3.89 $\downarrow$}} & 36.09 & \multirow{2}{*}{\shortstack{\\3.27 $\downarrow$}} & 33.35 & \multirow{2}{*}{\shortstack{\\2.81 $\downarrow$}} & 31.30 & \multirow{2}{*}{\shortstack{\\2.38 $\downarrow$}} & 29.62 & \multirow{2}{*}{\shortstack{\\2.11 $\downarrow$}} \\ 
        \arrayrulecolor{black!50}\cmidrule{1-1}\cmidrule{3-3}\cmidrule{5-5}\cmidrule{7-7}\cmidrule{9-9}\cmidrule{11-11}
        \cellcolor[gray]{0.9}\textbf{Ours} &  & \textbf{35.65} & & \textbf{32.82} & & \textbf{30.54} & & \textbf{28.92} & & \textbf{27.51} &\\ 
        \arrayrulecolor{black!100}\midrule
        TENT~\cite{wang2020tent} & \multirow{2}{*}{\shortstack{WRN-40-2}} & 35.53 &\multirow{2}{*}{\shortstack{\\2.82 $\downarrow$}}& 32.89 &\multirow{2}{*}{\shortstack{\\2.40 $\downarrow$}}& 30.72 &\multirow{2}{*}{\shortstack{\\1.87 $\downarrow$}}& 29.06 &\multirow{2}{*}{\shortstack{\\1.53 $\downarrow$}}& 27.67 &\multirow{2}{*}{\shortstack{\\1.29 $\downarrow$}}\\ 
        \arrayrulecolor{black!50}\cmidrule{1-1}\cmidrule{3-3}\cmidrule{5-5}\cmidrule{7-7}\cmidrule{9-9}\cmidrule{11-11}
        \cellcolor[gray]{0.9}\textbf{Ours} & & \textbf{32.71} & & \textbf{30.49} & & \textbf{28.85} & & \textbf{27.53} & & \textbf{26.38} & \\ 
        \arrayrulecolor{black!100}\bottomrule
        \end{tabular}
        }
    \label{tab_cifar100_all}
\end{subtable}
\label{tab_cifar}
\vspace{-0.35cm}
\end{table}

\vspace{-0.4cm}
\subsection{Robustness against Image Corruptions}
\vspace{-0.1cm}
Table~\ref{tab_cifar}(a) shows a comparison of the robustness between our method and recent test-time adaptation methods for the most severe corruptions on CIFAR-100-C. TFA~\cite{liu2021ttt++} and TTT\texttt{++}~\cite{liu2021ttt++} were originally implemented as offline adaptation methods that train a model by observing the same data multiple times across numerous epochs, so we change these methods to the online adaptation setting to reproduce the results.
Our proposed method significantly outperforms other state-of-the-art methods with large margins of 3.89$\%$ for ResNet-50 and 2.59$\%$ for WRN-40-2. Table~\ref{tab_cifar}(b) shows the results on the most severe corruptions of CIFAR-10-C. Our method consistently outperforms other methods on CIFAR-10/100-C datasets across various backbone networks. 
In particular, WRN-40-2, which is trained with AugMix~\cite{hendrycks2019augmix} for a data processing to increase the robustness of the model, outperforms the other backbone networks, and our method further enhances the performance by complementing it. 
Table~\ref{tab_cifar}(c) shows the results on CIFAR-100-C with all severity levels. Because severity denotes the strength of the corruption, it shows how much the distribution shift presents, and our method outperforms TENT~\cite{wang2020tent} at all levels with a large margin.

\begin{table*}[t!]
\caption{Ablation study on CIFAR-100-C. ResNet-50 is used.}
\vspace*{-0.2cm}
\begin{center}
\setlength\tabcolsep{5.5pt}
\renewcommand{\arraystretch}{0.9}
\resizebox{295pt}{!}{
\label{tab_ablation}
\begin{tabular}{l|c|c}
\toprule
\multicolumn{1}{c|}{\textbf{Methods}} & \textbf{Learning rate} & \textbf{Average err ($\%$)} \\
\drule
Main & 1e-3 & 75.70 \\
\arrayrulecolor{black!50}\midrule
Main (optimal learning rate) & 1e-4 & 39.44 \\ 
\midrule
Main + \textbf{NSP} $\left(\mathcal{L}^{\text{aux\_ent}}_{\theta_e}+\mathcal{L}^{\text{aux\_sel}}_{\theta_e}\right)$ & 1e-4 & 37.55 \\ 
\midrule
Main + \textbf{SWR} & 1e-3 & 37.76 \\ 
\midrule
Main + \textbf{SWR} + \textbf{NSP} $\left(\mathcal{L}^{\text{aux\_ent}}_{\theta_e}\right)$ & 1e-3 & 35.80 \\ 
\midrule
\cellcolor[gray]{0.9}Main + \textbf{SWR} + \textbf{NSP} $\left(\mathcal{L}^{\text{aux\_ent}}_{\theta_e}+\mathcal{L}^{\text{aux\_sel}}_{\theta_e}\right)$ & 1e-3 & \textbf{35.65} \\ 
\arrayrulecolor{black!100}\bottomrule
\end{tabular}
}
\end{center}
\vspace*{-0.77cm}
\end{table*}

\vspace*{-0.35cm}
\subsection{Ablation Studies}
\vspace*{-0.05cm}
Table~\ref{tab_ablation} shows the effectiveness of our proposed shift-agnostic weight regularization (SWR) and nearest source prototype (NSP) classifier through ablation studies.
At a high learning rate, optimizing only the main task loss based on the entropy of the model prediction results in poor performance, but adjusting the learning rate reduces the error rate to 39.44$\%$. 
Adding the NSP to the main task loss leads to the performance improvement of 1.89\%, and including the SWR improves the performance by 1.68\% even at a high learning rate. 
Our method with both SWR and NSP achieves 35.65\% error rate with 3.79\% performance enhancement compared to using only the main task loss.

\begin{table}[t!]
\caption{Comparison of error rate ($\%$) according to changes in projector depth.}
\vspace*{-0.2cm}
\begin{center}
\setlength\tabcolsep{5pt}
\renewcommand{\arraystretch}{0.9}
\footnotesize
\label{tab_projection_layer}
\resizebox{220pt}{!}{
\begin{tabular}{c|c|c|c|c|c}
\toprule
\multirow{2}{*}{\shortstack{\\ \\ \textbf{Datasets}}} & \multirow{2}{*}{\shortstack{\\ \\ \textbf{Backbone}}} & \multicolumn{4}{c}{\textbf{Projector depth}} \\
\cmidrule{3-6}
& & \cellcolor[gray]{0.9}\textbf{None} & \textbf{1} & \cellcolor[gray]{0.9}\textbf{2} & \textbf{3} \\
\drule
\multirow{2}{*}{\shortstack{\\ \\CIFAR-100-C}} & WRN-40-2 & 33.04 & 32.79 & \textbf{32.71} & 32.89 \\ 
\arrayrulecolor{black!50}\cmidrule{2-6}
 & ResNet-50 & 36.34 & \textbf{35.43} & 35.65 & 36.81     \\ 
\arrayrulecolor{black!100}\midrule
\multirow{4}{*}{CIFAR-10-C} & WRN-40-2 & \textbf{10.37} & 10.52 & 10.42 & 10.46 \\ 
\cmidrule{2-6}
 & WRN-28-10 & \textbf{15.70} & 16.45 & 16.09 & 16.39     \\ 
\arrayrulecolor{black!50}\cmidrule{2-6}
 & ResNet-50 & \textbf{12.52} & 12.91 & 12.95 & 12.87     \\ 
\arrayrulecolor{black!100}\bottomrule
\end{tabular}}
\end{center}
\vspace*{-0.75cm}
\end{table}

\vspace*{-0.39cm}
\subsection{Projector Design and Hyper-parameter Impacts}
\vspace*{-0.05cm}
Table~\ref{tab_projection_layer} shows the performance impact of changing the projector depth (\textit{i.e.,} number of projection layer). In addition, we conduct experiments to apply the auxiliary task loss $\mathcal{L}^{\text{aux}}_{\theta_e}$ directly to the feature representation $\boldsymbol{h}$, the encoder's output without using the projector. 
The model with the projector outperforms the one without the projector  on CIFAR-100-C, and opposite results are obtained on CIFAR-10-C.
Since the auxiliary task loss is applied to the embedding space based on the cosine similarity between the source prototypes and the target embeddings, its effect may be minimal if they are severely misaligned.
To compensate for this issue, we attach and train the projector that minimizes the misalignment between the source and target embeddings by enabling transformation-invariant mapping and bringing the projections belonging to the same class closer together in the new embedding space. However, if the number of classes is small (\textit{e.g.,} CIFAR-10-C), the source and target may already be relatively well aligned compared to the case with a large number of classes (\textit{e.g.,} CIFAR-100-C). In this case, we conjecture that applying the auxiliary task loss directly to the encoder's output $\boldsymbol{h}$ rather than the new embedding space $\boldsymbol{z}$, the projector's output, generates a better-aligned representation $\boldsymbol{h}$ between the source and target, which can be more helpful to the classifier.

Table~\ref{tab_projector} shows the experimental results according to (a) the projector width (\textit{i.e.,} output dimension of the last layer), (b) the transformation used for training the projector, and (c) whether to fine-tune or freeze the projector during test-time adaptation. Our default settings are marked with gray-colored cells, and these settings are also applied to the domain generalization benchmarks in the following section without additional tuning.

\begin{table}[t]
\caption{Hyper-parameter impacts on CIFAR-100-C. ResNet-50 is used.}
\vspace*{-0.2cm}
\begin{subtable}[h]{0.27\textwidth}
    \centering
    \setlength\tabcolsep{6pt}
    \renewcommand{\arraystretch}{0.7}
    \caption{}
    \vspace*{-0.15cm}
    \resizebox{86pt}{!}{
    \begin{tabular}{c|c}
    \toprule
    \textbf{Width} & \textbf{Error ($\%$)} \\
    \drule
    128 & 36.34 \\ 
    \arrayrulecolor{black!50}\midrule
    256 & 35.85  \\ 
    \midrule
    \cellcolor[gray]{0.9}\textbf{512} & \textbf{35.65}  \\ 
    \midrule
    1024 & 36.09  \\ 
    \arrayrulecolor{black!100}\bottomrule
    \end{tabular}
    \label{tab_projection_dims}
    }
\end{subtable}
\hfill
\begin{subtable}[h]{0.34\textwidth}
    \centering
    \setlength\tabcolsep{3pt}
    \renewcommand{\arraystretch}{1.0}
    \caption{}
    \vspace*{-0.15cm}
    \resizebox{120pt}{!}{
    \begin{tabular}{l|c}
    \toprule
    \multicolumn{1}{c|}{\textbf{Transformation}} & \textbf{Error ($\%$)} \\
    \drule
    No transform & 37.88 \\ 
    \arrayrulecolor{black!50}\midrule
    Color distortion & 35.79 \\ 
    \midrule
    \cellcolor[gray]{0.9}\; \textbf{$+$Crop. \& Blur.} & \textbf{35.65} \\ 
    \arrayrulecolor{black!100}\bottomrule
    \end{tabular}
    \label{tab_projection_aug}
    }
\end{subtable}
\hfill
\begin{subtable}[h]{0.36\textwidth}
    \centering
    \setlength\tabcolsep{4pt}
    \renewcommand{\arraystretch}{1.1}
    \caption{}
    \vspace*{-0.15cm}
    \resizebox{128pt}{!}{
    \begin{tabular}{c|c|c}
    \toprule
    \multirow{2}{*}{\shortstack{\\ \\ \textbf{Backbone}}} & \multicolumn{2}{c}{\textbf{Error ($\%$)}} \\
    \arrayrulecolor{black!50}\cmidrule{2-3}
     & WRN-40-2 & ResNet-50 \\
    \arrayrulecolor{black!100}\drule
    \cellcolor[gray]{0.9}\textbf{Freeze.} & \textbf{32.71} & \textbf{35.65} \\ 
    \arrayrulecolor{black!50}\midrule
    \textbf{Finetune.} & 32.85 & 35.96 \\ 
    \arrayrulecolor{black!100}\bottomrule
    \end{tabular}
    \label{tab_projection_dims}
    }
\end{subtable}
\label{tab_projector}
\vspace*{-0.3cm}
\end{table}


\vspace*{-0.35cm}
\subsection{Quick Adaptation}\label{sec:exp_quick_adapt}
\vspace*{-0.05cm}
As shown in Table~\ref{tab_quick_adapt}, it is natural that the supervised method performs perfectly when learning and evaluating the same test samples iteratively. However, interestingly our method outperforms the supervised one in an online setting where the test sample is seen only once. Unlike the other methods that require a low learning rate to train (Fig.~\ref{fig:intro}(b),(d)), our method updates the entire parameters at a high learning rate. We conjecture that SWR enables quick convergence without performance degradation because only parameters sensitive to distribution shift (\textit{i.e.,} parameters that need to quickly adapt to a new domain) are largely updated with a high learning rate.

\begin{table}[t]
\vspace*{-0.0cm}
\caption{Comparison of error rate ($\%$) on CIFAR-100-C. Our method outperforms the supervised method in an online setting. LR denotes a learning rate.}
\vspace*{-0.2cm}
\begin{center}
\setlength\tabcolsep{5pt}
\renewcommand{\arraystretch}{0.8}
\footnotesize
\label{tab_quick_adapt}
\resizebox{300pt}{!}{
\begin{tabular}{c|c|c|c|c|c}
\toprule
\multirow{2}{*}{\shortstack{\\ \\ \textbf{Methods}}} & \multirow{2}{*}{\shortstack{\\ \textbf{GT}\\ \textbf{Label}}} & \multirow{2}{*}{\shortstack{\\\textbf{Optimal}\\ \textbf{LR}}} & \multicolumn{3}{c}{\textbf{Epoch}} \\
\arrayrulecolor{black!50}\cmidrule{4-6}
& & & \textbf{1} (online) & \textbf{2} (offline) & \textbf{3} (offline) \\
\arrayrulecolor{black!100}\drule
Entropy Minimization & No & 1e-4 & 39.81 & 38.84 & 39.08 \\ 
\arrayrulecolor{black!50}\midrule
Cross Entropy (Supervised) & Yes & 2e-4 & 38.27 & \textbf{7.41} & \textbf{0.86} \\ 
\arrayrulecolor{black!50}\midrule
\cellcolor[gray]{0.9}\textbf{Ours} & No & 1e-3 & \textbf{35.65} & 33.34 & 33.25 \\ 
\arrayrulecolor{black!100}\bottomrule
\end{tabular}}
\end{center}
\vspace*{-0.6cm}
\end{table}

\vspace*{-0.35cm}
\subsection{Domain Generalization Benchmarks}
\vspace*{-0.05cm}
To evaluate our method on the DG benchmarks, we follow the protocol proposed by DomainBed~\cite{gulrajani2020search} and T3A~\cite{iwasawa2021test}. Our method is model-agnostic, so we apply it to the pre-trained models using empirical risk minimization (ERM)~\cite{vapnik1999overview} or CORAL~\cite{sun2016deep} on the source domain in order to adapt the models to the target domain at test time. We use the leave-one-domain-out validation~\cite{gulrajani2020search} for model selection in all experiments in Table~\ref{tab_domain}. Our methods show state-of-the-art performance on average over four datasets and especially outperform T3A~\cite{iwasawa2021test} and the source pre-trained models with a large margin on PACS, OfficeHome, and TerraIncognita datasets. The detailed experimental setup can be found in the supplementary Section C.


\begin{table}[t!]
\caption{Comparison of accuracy ($\%$) on four DG benchmarks. $^\dagger$ denotes the reported results from DomainBed~\cite{gulrajani2020search}, and the others are reproduced values.}
\vspace*{-0.18cm}
\begin{center}
\setlength\tabcolsep{7pt}
\footnotesize
\label{tab_domain}
\resizebox{320pt}{!}{
\begin{tabular}{l|c|c|c|c|c}
\toprule
\multicolumn{1}{c|}{\textbf{Methods}} & \textbf{VLCS} & \textbf{PACS} & \textbf{OfficeHome} & \textbf{Terra} & \textbf{Average} \\
\drule
ERM$^\dagger$ & 76.8$\pm$1.0 & 83.3$\pm$0.6 & 67.3$\pm$0.3 & 46.2$\pm$0.2  & 68.4 \\ 
\arrayrulecolor{black!20}\cmidrule{1-5}
CORAL$^\dagger$ & \textbf{77.0}$\pm$0.5 & \textbf{83.6}$\pm$0.6 & \textbf{68.6}$\pm$0.2 & \textbf{48.1}$\pm$1.3 & 69.3 \\ 
\arrayrulecolor{black!100}\cmidrule{1-5}
\cmidrule{1-5}
ERM & 77.4$\pm$0.9 & 83.5$\pm$0.7 & 65.6$\pm$0.4 & 47.1$\pm$1.1 & 68.4 \\ 
\;\;$+$T3A & \textbf{79.4}$\pm$0.4 & 86.5$\pm$0.3 & 67.8$\pm$0.5 & 45.6$\pm$0.7 & 69.8 \\ 
\cellcolor[gray]{0.9}\;\;$+$\textbf{Ours} & 77.0$\pm$0.5 & \textbf{88.9}$\pm$0.1 & \textbf{69.2}$\pm$0.1 & \textbf{49.5}$\pm$0.8 & \textbf{71.2} \\ 
\arrayrulecolor{black!20}\cmidrule{1-5}
CORAL & 77.9$\pm$0.9 & 85.3$\pm$0.1 & 67.8$\pm$0.3 & 44.1$\pm$0.4 & 68.8 \\ 
\;\;$+$T3A & \textbf{79.3}$\pm$0.3 & 86.3$\pm$0.2 & 69.5$\pm$0.2 & 45.4$\pm$1.2 & 70.1 \\ 
\cellcolor[gray]{0.9}\;\;$+$\textbf{Ours} & 78.7$\pm$0.4 & \textbf{89.9}$\pm$0.1 & \textbf{71.0}$\pm$0.0 & \textbf{47.5}$\pm$0.6 & \textbf{71.8} \\ 
\arrayrulecolor{black!100}\bottomrule
\end{tabular}}
\end{center}
\vspace*{-0.2cm}
\end{table}

\vspace*{-0.35cm}
\subsection{Qualitative Results}
\begin{figure}[!t]
\vspace*{-0.10cm}
\centering
\includegraphics[width=0.99\linewidth]{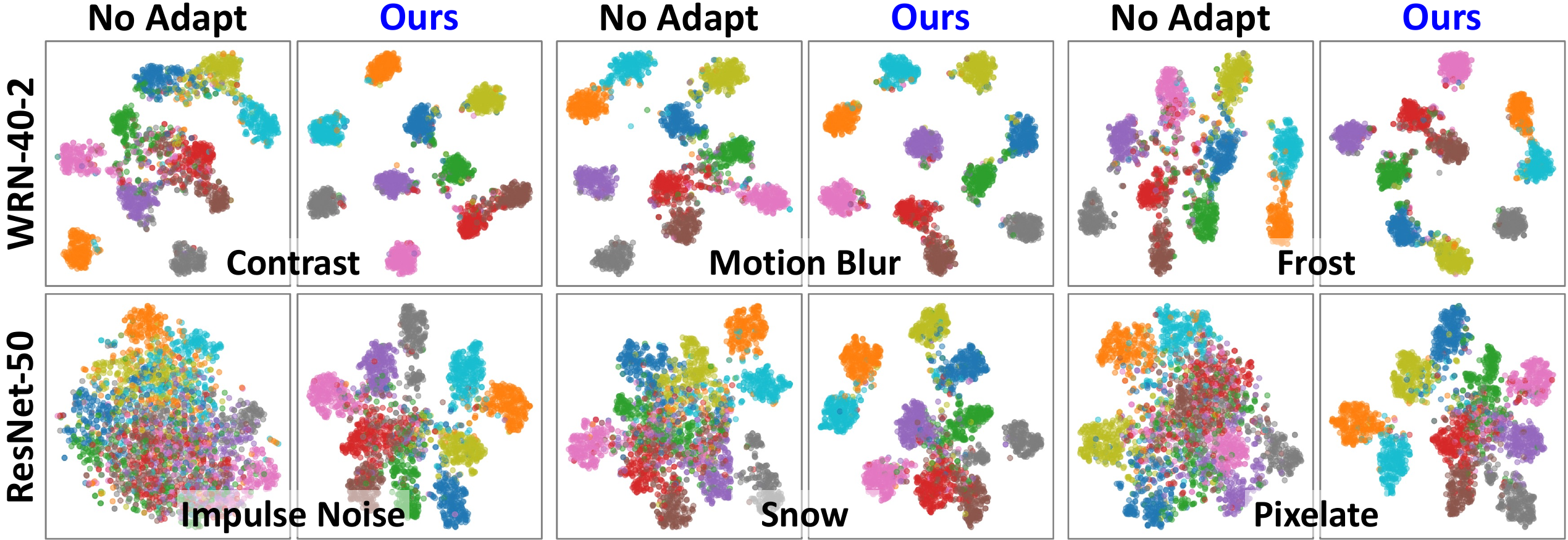}
\vspace*{-0.18cm}
\caption{t-SNE visualization of features from the target domain (CIFAR-10-C).}
\label{fig:exp_tsne}
\vspace*{-0.4cm}
\end{figure}

\vspace*{-0.05cm}
Fig.~\ref{fig:exp_tsne} visualizes the features on CIFAR-10-C using t-SNE~\cite{van2008visualizing}. The results in the first row are from WRN-40-2 as a source pre-trained model, and the results in the second row are from ResNet-50.
Even without test-time adaptation, WRN-40-2 (AugMix)~\cite{hendrycks2019augmix} is more robust against corruptions than ResNet-50, so better results can be obtained.
Our method significantly improves the performance in terms of intra-class cohesion and inter-class separation in both backbones.

\vspace*{-0.32cm}
\section{Conclusions}
\vspace*{-0.15cm}
This paper proposed two novel approaches for model-agnostic test-time adaptation. Our proposed shift-agnostic weight regularization enables the model to reliably and quickly adapt to unlabeled online data from the target domain by controlling the update of the model parameters according to their sensitivity to the distribution shift. In addition, our proposed auxiliary task based on the nearest source prototype classifier boosts the performance by aligning the source and target representations. Test-time adaptation is a challenging but promising area in terms of allowing the model to evolve itself while adapting to a new environment without human intervention. In this regard, our efforts aim to promote the importance of this field and stimulate new research directions.

\vspace*{-0.3cm}
\subsubsection{Acknowledgement} We would like to thank Kyuwoong Hwang, Simyung Chang, Hyunsin Park, Juntae Lee, Janghoon Cho, Hyoungwoo Park, Byeonggeun Kim, and Hyesu Lim of the Qualcomm AI Research team for their valuable discussions.

\bibliographystyle{style/splncs04}
\bibliography{egbib}
\clearpage

\title{Supplementary Material on\\Improving Test-Time Adaptation via Shift-agnostic Weight Regularization and Nearest Source Prototypes} 
\titlerunning{SWR \& NSP}
\author{Sungha Choi\thanks{Corresponding author.}\;\,
Seunghan Yang\;\,
Seokeon Choi\;\,
Sungrack Yun\\
}

\institute{Qualcomm AI Research\thanks{Qualcomm AI Research is an initiative of Qualcomm Technologies, Inc.}\\
\footnotesize{\vspace{0.0cm}\email{\{sunghac,seunghan,seokchoi,sungrack\}@qti.qualcomm.com}}}
\authorrunning{S. Choi et al.}

\maketitle
\setcounter{table}{6}
\setcounter{figure}{6}
\setcounter{equation}{9}

\def\thesection{\Alph{section}}
\setcounter{section}{0}

This supplementary material begins with a discussion of our two proposed approaches and then provides additional quantitative results examining hyper-parameter impacts, the performance of the NSP classifier, the transform function of the SWR, and the SWR variants. We also evaluate our approach on a large-scale dataset, ImageNet-C~\cite{hendrycks2018benchmarking}, and expand the proposed SWR to support pixel-level classification (\textit{i.e.,} semantic segmentation) on Cityscapes-C~\cite{Cordts2016Cityscapes,hendrycks2018benchmarking}. Finally, we provide further implementation details for the projector and additional information about the experiments on the domain generalization benchmarks.

\section{Discussion} \label{supp_discuss}
\subsection{Shift-agnostic Weight Regularization}
The intuition of SWR is to control the update of model parameters depending on each parameter's sensitivity to the distribution shift. If $\boldsymbol{\theta}^*$ in Eq.~(\textcolor{red}{1}) is fixed as source model parameters without being updated with the model parameters from the previous step, it is difficult to adapt the model to the target data sufficiently. Constraining the model parameters not to move away significantly from the source model parameters is not the purpose of SWR (Instead, NSP aligns source and target features based on the source prototypes). Updating $\boldsymbol{\theta}^*$ shows better performance than freezing $\boldsymbol{\theta}^*$, and these results are included in the supplementary Section~\ref{supp_freezing_theta}.

We assumed color and blur as a distribution shift to find shift-agnostic and shift-biased weights. Using too various augmentations increases the number of shift-biased weights. This means that more parameters are largely updated, which may result in performance degradation (Table~\ref{tab_supp_transform}). We conjecture that the augmentations such as color and blur, which can be commonly included in various domain gaps, are suitable for finding shift-agnostic weights.

We generated the penalty vector $\boldsymbol{w}$ in parameter-wise, output channel-wise, and layer\footnote{torch.nn.Module units defined in Pytorch.}-wise manners, and we chose the best method experimentally. While SpotTune~\cite{guo2019spottune} considers residual block units (16 units for ResNet-50) in terms of where to fine-tune, we generate the penalty values on layer$^{\textcolor{red}{1}}$ units (161 units for ResNet-50), which is much more granular than SpotTune.

\subsection{Superiority of Nearest Source Prototypes}%
The purpose of the NSP is twofold: (1) aligning target and source features by leveraging the source prototypes as reference points (Fig.~\textcolor{red}{5}(b), $\mathcal{L}^{\text{aux\_ent}}_{\theta_e}$), and (2) learning input consistency (Fig.~\textcolor{red}{5}(c), $\mathcal{L}^{\text{aux\_sel}}_{\theta_e}$). 
As shown in Table~\textcolor{red}{2}, $\mathcal{L}^{\text{aux\_ent}}_{\theta_e}$ has more crucial contribution than $\mathcal{L}^{\text{aux\_sel}}_{\theta_e}$. To further validate the superiority of NSP, we conduct experiments while keeping SWR but replacing NSP with FixMatch~\cite{sohn2020fixmatch} using this Pytorch implementation\footnote{https://github.com/kekmodel/FixMatch-pytorch/blob/master/train.py} on settings (a) and (b) in Table~\textcolor{red}{1}. FixMatch improves performance by up to 0.13\% compared to using SWR alone (up to 2.11\% increase when NSP is applied). We can find that learning only input consistency, such as applying FixMatch, is not sufficient to handle the distribution shift between source and target.

\section{Further Experiments} \label{supple}

\subsection{Impact of Number of Source Samples in SWR}

\begin{figure}[!t]
\centering
\includegraphics[width=0.74\linewidth]{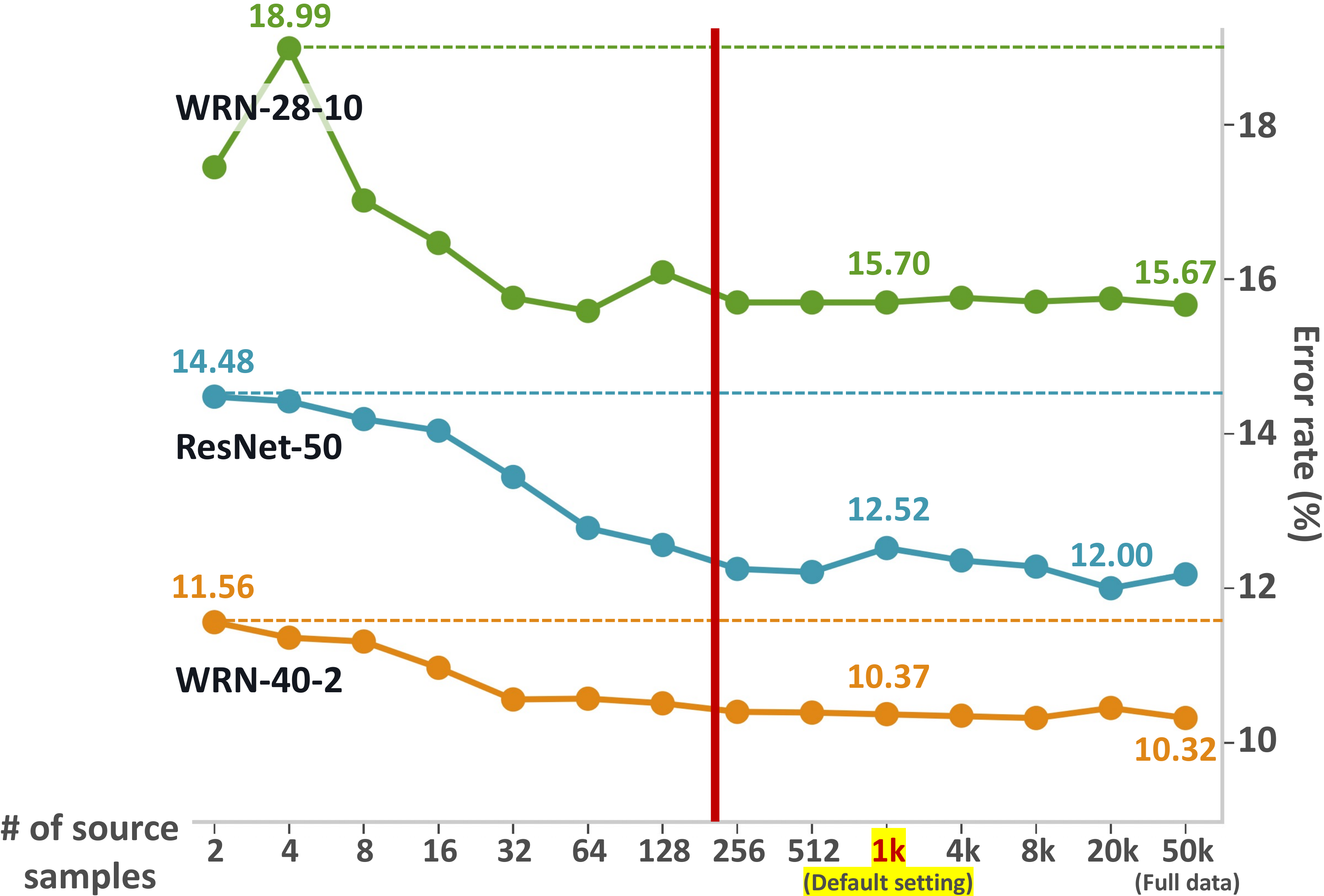}
\vspace{-0.0cm}
\caption{Comparison of error rate ($\%$) according to the number of source samples (\textit{i.e.,} the images in CIFAR-10~\cite{krizhevsky2009learning} train set) used to obtain the sensitivity of each model parameter to distribution shift in the SWR. The x- and y-axes denote the number of source samples and the error rate on CIFAR-10-C~\cite{hendrycks2018benchmarking}, respectively.}
\label{fig:supp_sample}
\vspace{-0.1cm}
\end{figure}
As described in Section~\textcolor{red}{3.1}, the penalty vector $\boldsymbol{w}$ is calculated by employing the average cosine similarity between two gradient vectors $\boldsymbol{g}$ and $\boldsymbol{g}'$ from $N$ source samples. 
We conduct the experiment to analyze the impact of the number of source samples on performance. As shown in Fig.~\ref{fig:supp_sample}, the smaller the number $N$ of source samples, the higher the error rate. However, with more than 256 images, performance tends to remain stable and low regardless of the number of source samples. We use 1k source samples as the default setting for all experiments, as mentioned in Section~\textcolor{red}{4.2}. This experiment shows that not all source data is required for the SWR.

\begin{table}[t!]
\vspace*{-0.0cm}
\caption{Comparison of error rate ($\%$) according to the changes in importance of each loss term. Gray-colored cells denote the default value. $H(\bar{p})\downarrow$ denotes entropy minimization, and $H(\bar{p})\uparrow$ indicates mean entropy maximization.}
\vspace*{-0.1cm}
\begin{center}
\setlength\tabcolsep{9pt}
\renewcommand{\arraystretch}{0.78}
\footnotesize
\label{tab_supp_lambda}
\resizebox{280pt}{!}{
\begin{tabular}{c|c|c|c|c|c|c}
\toprule
\multicolumn{2}{c|}{$H(p)\downarrow$} & \multicolumn{2}{c|}{$H(\bar{p})\uparrow$} & \textbf{SSL} & \textbf{Reg.} & \multirow{2}{*}{\shortstack{\\ \\ \textbf{Err.}}} \\
\arrayrulecolor{black!50}\cmidrule{1-6}
$\;\lambda_{m_1}\;$ & $\;\lambda_{a_1}\;$ & $\;\lambda_{m_2}\;$ & $\;\lambda_{a_2}\;$ & $\quad\lambda_s\quad$ & $\quad\lambda_r\quad$ &    \\
\arrayrulecolor{black!100}\drule
\cellcolor[gray]{0.9}0.2 & \cellcolor[gray]{0.9}0.8 & \cellcolor[gray]{0.9}0.25 & \cellcolor[gray]{0.9}0.25 & \cellcolor[gray]{0.9}0.1 & \cellcolor[gray]{0.9}250 & \textbf{35.65} \\ 
0.5 & 0.5 & \cellcolor[gray]{0.9}0.25 & \cellcolor[gray]{0.9}0.25 & \cellcolor[gray]{0.9}0.1 & \cellcolor[gray]{0.9}250 & 36.05 \\ 
0.8 & 0.2 & \cellcolor[gray]{0.9}0.25 & \cellcolor[gray]{0.9}0.25 & \cellcolor[gray]{0.9}0.1 & \cellcolor[gray]{0.9}250 & 36.73 \\ 
\arrayrulecolor{black!100}\midrule
\cellcolor[gray]{0.9}0.2 & \cellcolor[gray]{0.9}0.8 & 0.1 & 0.1 & \cellcolor[gray]{0.9}0.1 & \cellcolor[gray]{0.9}250 & 35.85 \\ 
\cellcolor[gray]{0.9}0.2 & \cellcolor[gray]{0.9}0.8 & \cellcolor[gray]{0.9}0.25 & \cellcolor[gray]{0.9}0.25 & \cellcolor[gray]{0.9}0.1 & \cellcolor[gray]{0.9}250 & \textbf{35.65} \\ 
\cellcolor[gray]{0.9}0.2 & \cellcolor[gray]{0.9}0.8 & 0.5 & 0.5 & \cellcolor[gray]{0.9}0.1 & \cellcolor[gray]{0.9}250 & 35.92 \\ 
\arrayrulecolor{black!100}\midrule
\cellcolor[gray]{0.9}0.2 & \cellcolor[gray]{0.9}0.8 & \cellcolor[gray]{0.9}0.25 & \cellcolor[gray]{0.9}0.25 & 0.01 & \cellcolor[gray]{0.9}250 & 35.79 \\ 
\cellcolor[gray]{0.9}0.2 & \cellcolor[gray]{0.9}0.8 & \cellcolor[gray]{0.9}0.25 & \cellcolor[gray]{0.9}0.25 & \cellcolor[gray]{0.9}0.1 & \cellcolor[gray]{0.9}250 & \textbf{35.65} \\ 
\cellcolor[gray]{0.9}0.2 & \cellcolor[gray]{0.9}0.8 & \cellcolor[gray]{0.9}0.25 & \cellcolor[gray]{0.9}0.25 & 0.5 & \cellcolor[gray]{0.9}250 & 36.13 \\ 
\arrayrulecolor{black!100}\midrule
\cellcolor[gray]{0.9}0.2 & \cellcolor[gray]{0.9}0.8 & \cellcolor[gray]{0.9}0.25 & \cellcolor[gray]{0.9}0.25 & \cellcolor[gray]{0.9}0.1 & 10 & 53.01 \\ 
\cellcolor[gray]{0.9}0.2 & \cellcolor[gray]{0.9}0.8 & \cellcolor[gray]{0.9}0.25 & \cellcolor[gray]{0.9}0.25 & \cellcolor[gray]{0.9}0.1 & 100 & 36.78 \\ 
\cellcolor[gray]{0.9}0.2 & \cellcolor[gray]{0.9}0.8 & \cellcolor[gray]{0.9}0.25 & \cellcolor[gray]{0.9}0.25 & \cellcolor[gray]{0.9}0.1 & \cellcolor[gray]{0.9}250 & \textbf{35.65} \\ 
\cellcolor[gray]{0.9}0.2 & \cellcolor[gray]{0.9}0.8 & \cellcolor[gray]{0.9}0.25 & \cellcolor[gray]{0.9}0.25 & \cellcolor[gray]{0.9}0.1 & 500 & 36.29 \\ 
\cellcolor[gray]{0.9}0.2 & \cellcolor[gray]{0.9}0.8 & \cellcolor[gray]{0.9}0.25 & \cellcolor[gray]{0.9}0.25 & \cellcolor[gray]{0.9}0.1 & 1000 & 37.43 \\ 
\arrayrulecolor{black!100}\bottomrule
\end{tabular}}
\end{center}
\vspace*{-0.45cm}
\end{table}

\vspace*{-0.25cm}
\subsection{Importance of Each Loss Term}
\vspace*{-0.0cm}
As described in Section~\textcolor{red}{3}, our proposed losses are defined as

\vspace*{-0.1cm}
\begin{equation}
\begin{split}
\vspace{-0.1cm}
\mathcal{L}^{\text{target}}_{\theta_e,\theta_c} &=
\mathcal{L}^{\text{main}}_{\theta_e,\theta_c} +
\mathcal{L}^{\text{aux}}_{\theta_e} + 
\lambda_r\sum_l w_l\Vert\boldsymbol{\theta}_l-\boldsymbol{\theta}_l^*\Vert^2_2 \\ 
&=
\mathcal{L}^{\text{main}}_{\theta_e,\theta_c} +
\mathcal{L}^{\text{aux\_ent}}_{\theta_e} + \lambda_s\mathcal{L}^{\text{aux\_sel}}_{\theta_e} +
\lambda_r\sum_l w_l\Vert\boldsymbol{\theta}_l-\boldsymbol{\theta}_l^*\Vert^2_2 \\ 
&=
\begin{rcases}
\textcolor{blue!100}{\lambda_{m_1}}\frac{1}{N}\sum_{i=1}^N H(\tilde{y_i})-\textcolor{blue!100}{\lambda_{m_2}} H(\bar{\tilde{y}})\;\, \\
\end{rcases}\,\mathcal{L}^{\text{main}}_{\theta_e,\theta_c}\\
&\;\;+
\begin{rcases}
\textcolor{blue!100}{\lambda_{a_1}}\frac{1}{N}\sum_{i=1}^{N}H(\hat{y_i})-\textcolor{blue!100}{\lambda_{a_2}}H(\bar{\hat{y}})\;\;\\
\end{rcases}\,\mathcal{L}^{\text{aux\_ent}}_{\theta_e}\\
&\;\;+
\begin{rcases}
\textcolor{blue!100}{\lambda_s}\frac{1}{N}\sum_{i=1}^{N}\text{CE}\left(\hat{y}_i^k,{\hat{y}'^k_i}\right)\quad\quad\;\;\,\,\,\,\\
\end{rcases}\,\mathcal{L}^{\text{aux\_sel}}_{\theta_e}\\
&\;\;+\,\textcolor{blue!100}{\lambda_r}\sum_l w_l\Vert\boldsymbol{\theta}_l-\boldsymbol{\theta}_l^*\Vert^2_2, 
\vspace{-0.0cm}
\end{split}
\end{equation}
where $H(p)=-\sum_{k=1}^C p^k\log p^k$ and $\text{CE}\left(p,q\right)=-\sum_{k=1}^C p^k\log q^k$. Here, $\tilde{y}$ denotes the prediction of the main classifier, $\hat{y}$ is the prediction of the NSP classifier, symbol $\bar{\;}\,$ indicates average class probability distribution over batch samples, and symbol $\,'$ denotes the prediction of the transformed sample.
The hyper-parameters indicating the importance of each term are empirically set as $\lambda_{m_1}$=0.2, $\lambda_{a_1}$=0.8
$\lambda_{m_2}$=0.25, $\lambda_{a_2}$=0.25, $\lambda_s$=0.1, and $\lambda_r$=250. Table~\ref{tab_supp_lambda} shows the error rate (\%) according to the changes in importance of each term. The default settings are indicated by gray-colored cells.

\begin{table}[t!]
\caption{Comparison of error rate ($\%$) according to the combination of transform functions. We use the following transformations in Pytorch: ColorJitter (Color), RandomGrayscale (Gray), RandomInvert (Inve.), GaussianBlur (Blur), RandomHorizontalFlip (H.Fli.), and RandomResizedCrop (Crop.).}
\vspace*{-0.08cm}
\begin{center}
\setlength\tabcolsep{4pt}
\renewcommand{\arraystretch}{1.0}
\footnotesize
\label{tab_supp_transform}
\resizebox{330pt}{!}{
\begin{tabular}{c|c|c|c|c|c|c|c}
\toprule
\textbf{Datasets} &\textbf{Backbone} &
\textbf{$\,$Color$\,$} & \textbf{+Gray.} &
\textbf{+Inve.} 
&
\cellcolor[gray]{0.9}\textbf{+Blur.}
&
\textbf{+H.Fli.}
&
\textbf{+Crop.}  \\
\drule
\multirow{2}{*}{\shortstack{\\ \\CIFAR-100-C}} & WRN-40-2~\cite{zagoruyko2016wide,hendrycks2019augmix} & 33.23 & 33.04 & 32.99 & \textbf{32.71} & 32.79 & 33.31\\ 
\arrayrulecolor{black!30}\cmidrule{2-8}
 & ResNet-50~\cite{he2016deep} & 37.61 & 36.58 & 36.15 & \textbf{35.65} & 35.70 & 35.94      \\ 
\arrayrulecolor{black!100}\midrule
\multirow{4}{*}{CIFAR-10-C} & WRN-40-2 & 10.97 & 10.76 & 10.63 & 10.37 & \textbf{10.31} & 10.68 \\ 
\arrayrulecolor{black!30}\cmidrule{2-8}
 & WRN-28-10~\cite{zagoruyko2016wide} & 16.73 & 16.48 & 16.24 & \textbf{15.70} & 15.73 & 25.61 \\ 
\arrayrulecolor{black!30}\cmidrule{2-8}
 & ResNet-50~\cite{he2016deep} & 12.47 & 12.38 & 12.76 & 12.52 & 12.55 & \textbf{11.81}     \\ 
\arrayrulecolor{black!100}\bottomrule
\end{tabular}}
\end{center}
\vspace*{-0.45cm}
\end{table}

\vspace*{-0.23cm}
\subsection{Ablation Studies on Transform Functions of SWR}
\vspace*{-0.0cm}
As described in Section~\textcolor{red}{3.1}, the SWR employs the sensitivity of each model parameter to distribution shift, and we simulate the distribution shift through the transform functions such as color distortion and Gaussian blur. We conduct ablation studies on transform functions to find the optimal combination of transformations. Table~\ref{tab_supp_transform} shows the experimental results by adding each transformation in order. We use the following combinations as default setting: ColorJitter, RandomGrayscale, RandomInvert, and Gaussian Blur in Pytorch.
The pseudo-code for our default setting using Pytorch is as follows.

\footnotesize{
\begin{lstlisting}
from torchvision import transforms

TRANSFORMS_SWR = torch.nn.Sequential(
    transforms.ColorJitter(0.8, 0.8, 0.8, 0.2),
    RandomChoice(
        transforms.RandomGrayscale(p=0.5), 
        transforms.RandomInvert(p=0.5),
        p = 0.5
    ),
    RandomApply(
        transforms.GaussianBlur((3, 3), (1.0, 2.0)),
        p = 0.5
    ),
)
\end{lstlisting}
}

\begin{figure}[!t]
\centering
\includegraphics[width=\linewidth]{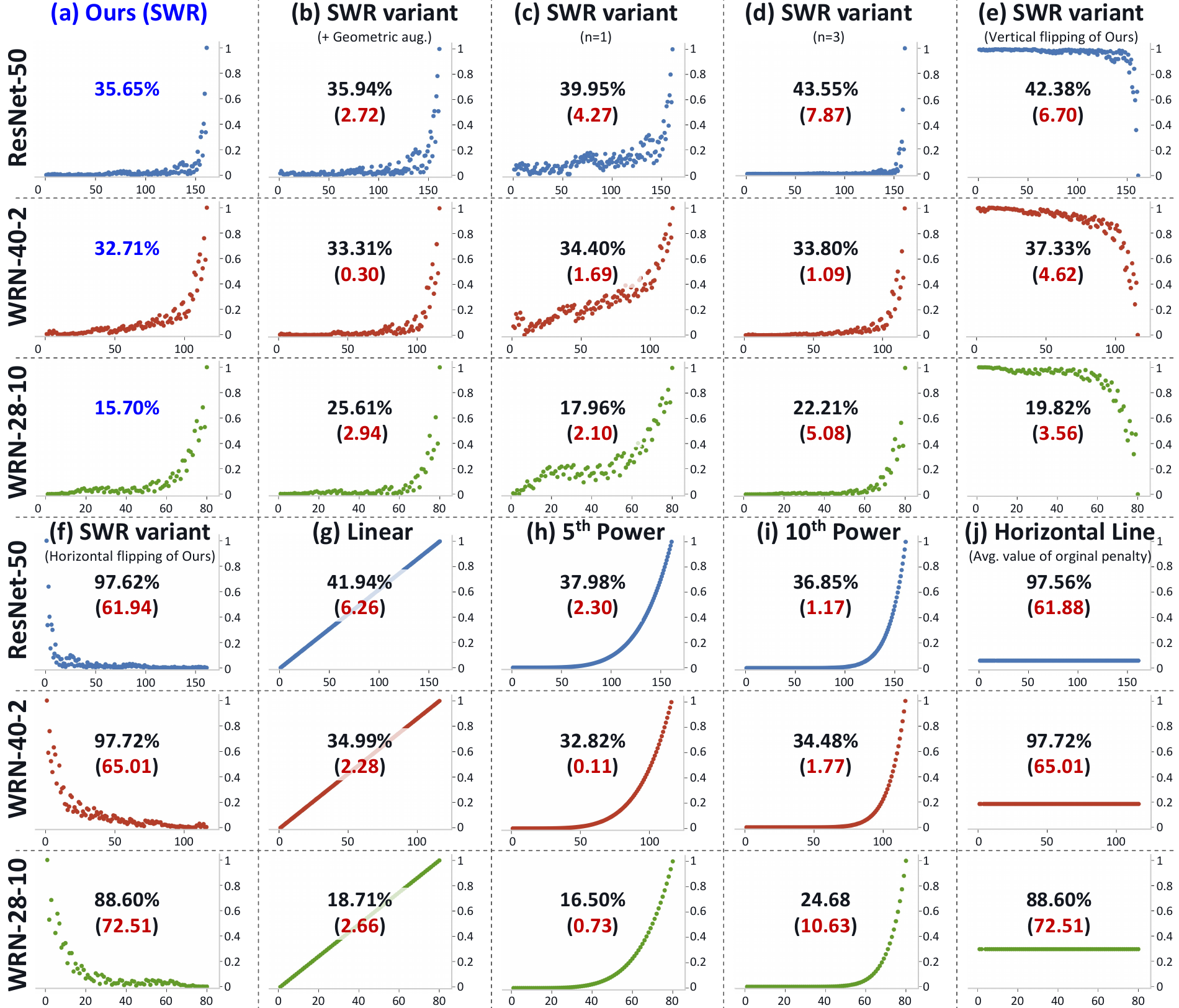}
\vspace*{-0.2cm}
\caption{Comparison of scatter plots of penalty vectors in various SWR variants. X- and y-axes indicate the layer index of the model and penalty value, respectively. Experiments using ResNet-50 and WRN-40-2 are conducted on CIFAR-100-C, and experiments with WRN-28-10 are performed on CIFAR-10-C. The number in each scatter plot indicates the error rate ($\%$), and the number in parentheses denotes the difference from the error rate of our proposed SWR.}
\label{fig:supp_swr}
\vspace*{-0.5cm}
\end{figure}

\vspace*{-0.1cm}
\subsection{Variations of Shift-agnostic Weight Regularization}
\vspace*{-0.0cm}
Fig.~\ref{fig:supp_swr} visualizes the penalty vector $\boldsymbol{w}$ in various SWR variants. X- and y-axes denote the layer index and penalty value, respectively. We can see that the penalty vector is different for each backbone network, and a high penalty is applied to the later layers. (b) is the result of using the combination of all transform functions listed in Table~\ref{tab_supp_transform}. (c) and (d) are the results of changing the exponent value of 2 in Eq. (\textcolor{red}{2}) into 1 and 3, respectively (\textit{i.e.,} $\boldsymbol{w}=\left(\nu\left\lbrack s_1,\dots,s_l,\dots,s_{L} \right\rbrack\right)^1$ and $\boldsymbol{w}=\left(\nu\left\lbrack s_1,\dots,s_l,\dots,s_{L} \right\rbrack\right)^3$). (e) and (f) are the results of flipping the original penalty vector of our proposed SWR vertically and horizontally, respectively. (g) to (j) are the results of employing manually designed functions without calculating the cosine similarity between two gradient vectors generated by back-propagation from the model's prediction of the source samples. Our proposed SWR outperforms the various variants of the SWR.

\begin{table}[t!]
\caption{Comparison of error rate ($\%$) between the main and NSP classifiers.}
\vspace*{-0.05cm}
\begin{center}
\setlength\tabcolsep{5pt}
\renewcommand{\arraystretch}{0.87}
\footnotesize
\label{tab_supp_nsp_classifier}
\resizebox{230pt}{!}{
\begin{tabular}{c|c|c|c}
\toprule
\multirow{2}{*}{\shortstack{\\ \\ \textbf{Datasets}}} & \multirow{2}{*}{\shortstack{\\ \\ \textbf{Backbone}}} & \multicolumn{2}{c}{\textbf{Classifier}} \\
\arrayrulecolor{black!30}\cmidrule{3-4}
& & \cellcolor[gray]{0.9}\textbf{\quad Main \quad} & \textbf{\quad NSP \quad} \\
\arrayrulecolor{black!100}\drule
\multirow{2}{*}{\shortstack{\\ \\CIFAR-100-C}} & WRN-40-2 & 32.71 & 35.48 \\ 
\arrayrulecolor{black!30}\cmidrule{2-4}
 & ResNet-50 & 35.65 & 36.52     \\ 
\arrayrulecolor{black!100}\midrule
\multirow{4}{*}{CIFAR-10-C} & WRN-40-2 & 10.37 & 10.41 \\ 
\arrayrulecolor{black!30}\cmidrule{2-4}
 & WRN-28-10 & 15.70 & 15.75 \\ 
\arrayrulecolor{black!30}\cmidrule{2-4}
 & ResNet-50 & 12.52 & 12.90 \\ 
\arrayrulecolor{black!100}\bottomrule
\end{tabular}}
\end{center}
\vspace*{-0.1cm}
\end{table}

\subsection{Performance of Nearest Source Prototype Classifier}
The goal of the NSP classifier is to improve the performance of the main classifier by aligning the source and target representations through its optimization, so the NSP classifier itself is not used for the classification. Table~\ref{tab_supp_nsp_classifier} shows the performance of the NSP classifier, and we can see that it is not as good as the performance of the main classifier.



\begin{table}[t!]
\vspace*{-0.2cm}
\caption{Comparison of error rate ($\%$) according to the methods of obtaining the source prototypes.}
\vspace*{-0.1cm}
\begin{center}
\setlength\tabcolsep{8pt}
\renewcommand{\arraystretch}{0.87}
\footnotesize
\label{tab_supp_prototype}
\resizebox{250pt}{!}{
\begin{tabular}{c|c|c|c|c}
\toprule
\multirow{2}{*}{\shortstack{\\ \\ \textbf{Datasets}}} & \multirow{2}{*}{\shortstack{\\ \\ \textbf{Backbone}}} &
\multicolumn{3}{c}{\textbf{Source prototypes}}  \\
\arrayrulecolor{black!30}\cmidrule{3-5}
& & \cellcolor[gray]{0.9}$\quad\boldsymbol{z}\quad$ & \cellcolor[gray]{0.9}$\quad\boldsymbol{h}\quad$ & $\quad\boldsymbol{\theta_c}\quad$  \\
\arrayrulecolor{black!100}\drule
\multirow{2}{*}{\shortstack{\\ \\CIFAR-100-C}} & WRN-40-2 & \textbf{32.71} & 33.04 & 34.48 \\ 
\arrayrulecolor{black!30}\cmidrule{2-5}
 & ResNet-50 & \textbf{35.65} & 36.34 & 39.89      \\ 
\arrayrulecolor{black!100}\midrule
\multirow{4}{*}{CIFAR-10-C} & WRN-40-2 & 10.42 & \textbf{10.37} & 11.62 \\ 
\arrayrulecolor{black!30}\cmidrule{2-5}
 & WRN-28-10 & 16.09 & 15.70 & \textbf{15.67}      \\ 
\arrayrulecolor{black!30}\cmidrule{2-5}
 & ResNet-50 & 12.95 & \textbf{12.52} & 14.74      \\ 
\arrayrulecolor{black!100}\bottomrule
\end{tabular}}
\end{center}
\vspace*{-0.75cm}
\end{table}

\begin{figure}[!t]
\centering
\includegraphics[width=\linewidth]{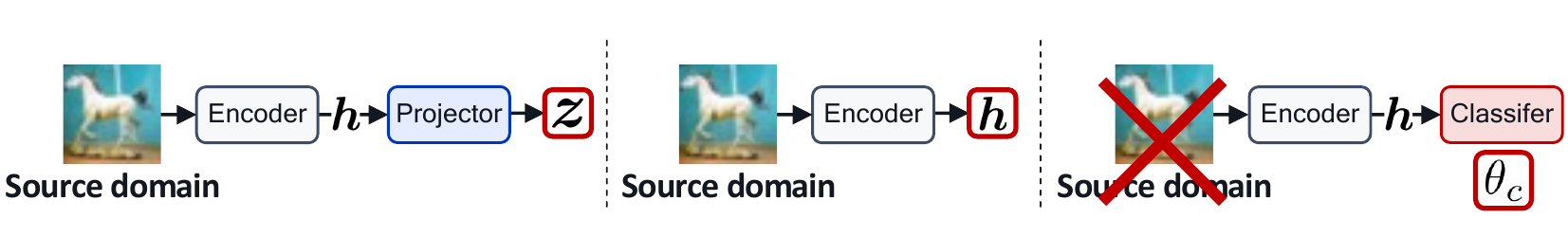}
\label{fig:supp_proto}
\vspace*{-0.2cm}
\end{figure}

\subsection{Source Prototypes}
Table~\ref{tab_supp_prototype} shows the performance comparison according to the methods of obtaining the source prototypes. The source prototypes can be generated and updated by an exponential moving average of projection $\boldsymbol{z}$ or feature representation $\boldsymbol{h}$ inferred across the source samples. Additionally, since the network weights for each class of the source pre-trained linear classifier can be considered as a source prototype, we report the result of employing the main classifier's parameter vectors $\boldsymbol{\theta_c}$ as source prototypes without forward-propagation of the source samples.

\begin{table}[t!]
\caption{Comparison of error rate ($\%$) between updating $\boldsymbol{\theta}^*$ and freezing $\boldsymbol{\theta}^*$ in shift-agnostic regularization term.}
\vspace*{-0.05cm}
\begin{center}
\setlength\tabcolsep{5pt}
\renewcommand{\arraystretch}{0.9}
\footnotesize
\label{tab_supp_theta}
\resizebox{200pt}{!}{
\begin{tabular}{c|c|c|c}
\toprule
\textbf{Datasets} & \textbf{Backbone} & Updating $\boldsymbol{\theta}^*$ & Freezing $\boldsymbol{\theta}^*$  \\
\drule
\multirow{2}{*}{\shortstack{\\ \\CIFAR-100-C}} & WRN-40-2 & \textbf{32.71} & 33.49 \\ 
\arrayrulecolor{black!50}\cmidrule{2-4}
 & ResNet-50 & \textbf{35.65} & 37.00 \\ 
\arrayrulecolor{black!100}\midrule
\multirow{4}{*}{CIFAR-10-C} & WRN-40-2 & \textbf{10.37} & 10.87 \\ 
\cmidrule{2-4}
 & WRN-28-10 & \textbf{15.70} & 18.30 \\ 
\arrayrulecolor{black!50}\cmidrule{2-4}
 & ResNet-50 & \textbf{12.52} & 13.02 \\ 
\arrayrulecolor{black!100}\bottomrule
\end{tabular}}
\end{center}
\vspace*{-0.2cm}
\end{table}

\begin{table}[b!]
\vspace{-0.2cm}
\caption{Comparison with other methods on ImageNet-C~\cite{hendrycks2018benchmarking} dataset. These results are reproduced in our environment. Source denotes the source pre-trained model without test-time adaptation.} 
\vspace{-0.1cm}
\begin{subtable}[h]{\textwidth}
    \centering
        \setlength\tabcolsep{2pt}
        \renewcommand{\arraystretch}{0.8}
        \footnotesize
        \caption{Comparison of error rate ($\%$) on ImageNet-C with severity level 5}
        \vspace{-0.1cm}
        \resizebox{\textwidth}{!}{
        \begin{tabular}{c|c|c|ccccccccccccccc}
        \toprule
        \rule{0pt}{5pt}\textbf{Backbone} & \textbf{Methods} & \textbf{Avg. err} & \textbf{Gaus.} & \textbf{Shot} & \textbf{Impu.} & \textbf{Defo.} & \textbf{Glas.} & \textbf{Moti.} & \textbf{Zoom} & \textbf{Snow} & \textbf{Fros.} & \, \textbf{Fog} \, & \textbf{Brig.} & \textbf{Cont.} & \textbf{Elas.} & \textbf{Pixe.} & \textbf{Jpeg} \\
        \drule
        \multirow{3}{*}{\shortstack{\\ WRN-40-2\\(AugMix)\\~\cite{zagoruyko2016wide,hendrycks2019augmix}}} & Source & 74.32 & 89.8 & 84.9 & 89.3 & 78.7 & 86.5 & 75.5 & 66.6 & 78.3 & 72.8 & 77.1 & 42.4 & 87.0 & 74.5 & 58.3 & 53.2 \\
        \arrayrulecolor{black!50}\cmidrule{2-18}
        & TENT~\cite{wang2020tent} & 51.17 & 66.8 & 63.2 & 63.6 & 64.3 & 65.1 & 47.8 & 42.8 & 45.1 & 52.0 & 40.6 & 31.9 & 60.8 & 40.8 & 37.6 & 45.2  \\
        \cmidrule{2-18}
        & \cellcolor[gray]{0.9}\textbf{Ours} & \textbf{48.01} & \textbf{59.8} & \textbf{56.5} & \textbf{58.4} & \textbf{61.1} & \textbf{62.8} & \textbf{44.6} & \textbf{41.5} & \textbf{41.6} & \textbf{47.8} & \textbf{38.5} & \textbf{30.2} & \textbf{58.2} & \textbf{39.8} & \textbf{36.3} & \textbf{43.1}  \\
        \midrule
        \multirow{3}{*}{\shortstack{\\ \\ResNet-50 \\~\cite{he2016deep}}} & Source & 93.34 & 96.1 & 95.9 & 96.3 & 98.3 & 98.3 & 97.2 & 94.1 & 97.1 & 93.5 & 97.0 & 74.5 & 99.9 & 96.1 & 87.3 & 78.6 \\
        \arrayrulecolor{black!50}\cmidrule{2-18}
        & TENT~\cite{wang2020tent} & 66.56 & 84.5 & 81.9 & 79.6 & 86.3 & 88.7 & 69.5 & 55.2 & 56.7 & 69.0 & 46.8 & \textbf{35.4} & \textbf{98.1} & \textbf{48.4} & \textbf{45.4} & 53.0  \\
        \cmidrule{2-18}
        & \cellcolor[gray]{0.9}\textbf{Ours} & \textbf{64.41} & \textbf{78.1} & \textbf{75.8} & \textbf{76.3} & \textbf{85.4} & \textbf{87.6} & \textbf{62.6} & \textbf{55.1} & \textbf{52.8} & \textbf{64.4} & \textbf{46.5} & 36.2 & 99.2 & 48.5 & 45.8 & \textbf{51.9}  \\
        \arrayrulecolor{black!100}\bottomrule
        \end{tabular}}
    \label{tab_sup_imagenet_5}
\end{subtable}
\vfill
\begin{subtable}[h]{\textwidth}
    \centering
        \setlength\tabcolsep{6pt}
        \renewcommand{\arraystretch}{0.3}
        \vspace{-0.1cm}
        \footnotesize
        \caption{Comparison of error rate ($\%$) on ImageNet-C  with all severity levels}
        \vspace{-0.1cm}
        \resizebox{\textwidth}{!}{
        \label{tab_all_level}
        \begin{tabular}{c|c|c|c|c|c|c|c|c|c|c|c|c|c}
        \toprule
        \textbf{Backbone} & \textbf{Methods} & \multicolumn{2}{c|}{\textbf{Avg. err}}& \multicolumn{2}{c|}{\textbf{Lv.5}} & \multicolumn{2}{c|}{\textbf{Lv.4}} & \multicolumn{2}{c|}{\textbf{Lv.3}} & \multicolumn{2}{c|}{\textbf{Lv.2}} & \multicolumn{2}{c}{\textbf{Lv.1}} \\ 
        \arrayrulecolor{black!100}\midrule
        \multirow{2}{*}{\shortstack{WRN-40-2}} & TENT~\cite{wang2020tent} & 39.27 &\multirow{2}{*}{\shortstack{1.88 $\downarrow$}}& 51.17 &\multirow{2}{*}{\shortstack{3.16 $\downarrow$}}& 42.94 &\multirow{2}{*}{\shortstack{2.11 $\downarrow$}}& 37.36 &\multirow{2}{*}{\shortstack{1.65 $\downarrow$}}& 34.35 &\multirow{2}{*}{\shortstack{1.33 $\downarrow$}}& 30.54 &\multirow{2}{*}{\shortstack{1.18 $\downarrow$}}\\ 
        \arrayrulecolor{black!50}\cmidrule{2-2}\cmidrule{3-3}\cmidrule{5-5}\cmidrule{7-7}\cmidrule{9-9}\cmidrule{11-11}\cmidrule{13-13}
        & \cellcolor[gray]{0.9}\textbf{Ours} & \textbf{37.39} & & \textbf{48.01} & & \textbf{40.83} & & \textbf{35.71} & & \textbf{33.02} & & \textbf{29.36} & \\ 
        \midrule
        \multirow{2}{*}{\shortstack{ResNet-50}} & TENT~\cite{wang2020tent} & 47.93 &\multirow{2}{*}{\shortstack{0.54 $\downarrow$}} & 66.56 &\multirow{2}{*}{\shortstack{2.15 $\downarrow$}}& 54.06 &\multirow{2}{*}{\shortstack{2.05 $\downarrow$}}& 45.04 &\multirow{2}{*}{\shortstack{0.24 $\downarrow$}}& \textbf{39.82} &\multirow{2}{*}{\shortstack{0.57 $\uparrow$}}& \textbf{34.15} &\multirow{2}{*}{\shortstack{1.18 $\uparrow$}}\\ 
        \arrayrulecolor{black!50}\cmidrule{2-2}\cmidrule{3-3}\cmidrule{5-5}\cmidrule{7-7}\cmidrule{9-9}\cmidrule{11-11}\cmidrule{13-13}
        & \cellcolor[gray]{0.9}\textbf{Ours} & \textbf{47.39} & & \textbf{64.41} & & \textbf{52.01} & & \textbf{44.80} & & 40.39 & & 35.33 & \\ 
        \arrayrulecolor{black!100}\bottomrule
        \end{tabular}
        }
    \label{tab_sup_imagenet_all}
\end{subtable}
\label{tab_sup_imagenet}
\vspace{-0.0cm}
\end{table}

\subsection{Freezing $\boldsymbol{\theta}^*$ as source model parameters in SWR}\label{supp_freezing_theta}
Recall shift-agnostic weight regularization (SWR) term in Eq.~(\textcolor{red}{1}). We update $\boldsymbol{\theta}^*$ with the model parameters from the previous update step during the optimization trajectory. Alternatively, we can consider freezing $\boldsymbol{\theta}^*$ as a source pre-trained model and compare the performance between updating $\boldsymbol{\theta}^*$ and freezing $\boldsymbol{\theta}^*$, as shown in Table~\ref{tab_supp_theta}. Although freezing $\boldsymbol{\theta}^*$ performs worse than updating $\boldsymbol{\theta}^*$, the advantage of restricting the model parameters not to deviate significantly from the source pre-trained model (\textit{i.e.,} freezing $\boldsymbol{\theta}^*$) seems worth exploring in future work.

\begin{table}[t!]
\vspace*{-0.0cm}
\caption{Test-time adaptation (TTA) performance using SWR on Cityscapes-C dataset. DG denotes domain generalization.}
\vspace*{-0.12cm}
\begin{center}
\setlength\tabcolsep{6pt}
\resizebox{250pt}{!}{
\begin{tabular}{l|l}
\toprule
\multicolumn{1}{c|}{\textbf{Models (Cityscapes$\rightarrow$Cityscapes-C)}} & \textbf{mIoU} \\
\drule
DeepLabV3\texttt{+}~\cite{chen2018encoder} (ResNet-50) & 27.3\% \\ 
\cellcolor[gray]{0.9}\;\;$+$TTA (Main\texttt{+}SWR)  & \cellcolor[gray]{0.9}\textbf{49.8\%} ($\uparrow$ 22.5\%)\\ 
\arrayrulecolor{black!100}\midrule
RobustNet~\cite{choi2021robustnet} (ResNet-50+DG)  & 44.4\% \\ 
\cellcolor[gray]{0.9}\;\;$+$TTA (Main\texttt{+}SWR)  & \cellcolor[gray]{0.9}\textbf{55.2\%} ($\uparrow$ 10.8\%)\\ 
\arrayrulecolor{black!100}\bottomrule
\end{tabular}}
\end{center}
\label{supp_cityscapes}
\end{table}

\begin{table}[t!]
\vspace*{-0.55cm}
\caption{Comparison of performance with and without SWR according to the learning rate change. Test-time adaptation with the proposed SWR shows superior performance and less sensitivity to changes in the learning rate. LR denotes a learning rate.}
\vspace*{-0.12cm}
\begin{center}
\setlength\tabcolsep{6pt}
\resizebox{310pt}{!}{
\begin{tabular}{l|c|c|c}
\toprule
\multicolumn{1}{c|}{\textbf{Models (Cityscapes$\rightarrow$Cityscapes-C)}} & LR: 1e-4 & LR: 1e-5 & LR: 1e-6 \\
\drule
RobustNet~\cite{choi2021robustnet}$+$TTA (Main) & 7.3\% & 28.5\% & 53.1\% \\ 
\arrayrulecolor{black!20}\midrule
\cellcolor[gray]{0.9}RobustNet~\cite{choi2021robustnet}$+$TTA (Main\texttt{+}SWR) & \textbf{53.7\%} & \textbf{55.2\%} & \textbf{54.0\%} \\ 
\arrayrulecolor{black!100}\bottomrule
\end{tabular}}
\end{center}
\label{supp_cityscapes_lr}
\vspace*{-0.35cm}
\end{table}

\subsection{Experiments on ImageNet-C}
We further validate our method on ImageNet-C~\cite{hendrycks2018benchmarking} dataset.
Following the ImageNet-C experiment in TENT~\cite{wang2020tent}, the batch size and learning rate are set to 64 and 0.00025, respectively. We only change the importance of the regularization term, $\lambda_r$, to 3000 from the default value of the hyper-parameters specified in Section~\textcolor{red}{4.2} for experiments in this section. Table~\ref{tab_sup_imagenet} demonstrates that our method is superior to TENT~\cite{wang2020tent}.

\subsection{Experiments on Cityscapes-C}
Although this paper has focused on the image classification task, we further demonstrate the scalability of our proposed method by expanding it to support the pixel-level classification (\textit{i.e.,} semantic segmentation).
We use two ResNet-50-based pre-trained models, DeepLabV3\texttt{+}~\cite{chen2018encoder} and RobustNet~\cite{choi2021robustnet}, as baseline models for test-time adaptation. RobustNet can be a better starting point for test-time adaptation as it is more robust to distribution shift than DeepLabV3 due to improved domain generalization.
We consider the original Cityscapes~\cite{Cordts2016Cityscapes} dataset to be the source data and generate Cityscapes-C dataset by applying algorithmically created corruption~\cite{hendrycks2018benchmarking} to the original Cityscapes. The Cityscapes-C dataset is regarded as unlabeled online test data.

We adapt the Cityscapes pre-trained models to the Cityscapes-C using the loss as
\vspace{-0.0cm}
\begin{equation}
\mathcal{L}^{\text{target}}_{\theta_e,\theta_c} =
\mathcal{L}^{\text{main}}_{\theta_e,\theta_c} +
\lambda_r\sum_l w_l\Vert\boldsymbol{\theta}_l-\boldsymbol{\theta}_l^*\Vert^2_2.
\vspace{-0.0cm}
\end{equation}
Note that the loss includes only SWR, not NSP. The integration of NSP into the test-time adaptation loss for semantic segmentation is left for future work. The batch size, learning rate, and the importance $\lambda_r$ of the regularization term are set to 2, 1e-5, and 40, respectively. In contrast to the test-time adaptation on the image classification task, which discards running estimates and uses batch statistics on test data, the running statistics of batch normalization layers are kept and updated on test data with a momentum of 0.1 during test time, and we set the $\theta^*$ as the parameters of the source pre-trained model.

\begin{figure}[!t]
\centering
\includegraphics[width=\linewidth]{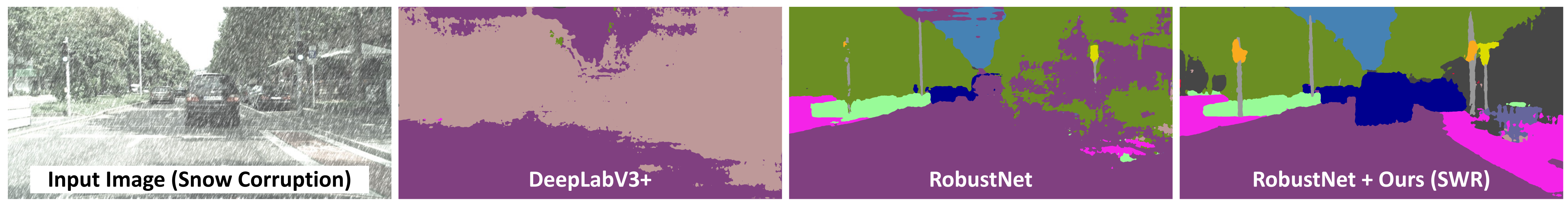}
\vspace{-0.45cm}
\caption{Comparison of segmentation results on Cityscapes-C~\cite{Cordts2016Cityscapes,hendrycks2018benchmarking} (snow corruption) between DeepLabV3+~\cite{chen2018encoder}, RobustNet~\cite{choi2021robustnet}, and ours}
\label{fig:supp_cityscapes}
\vspace{-0.0cm}
\end{figure}

Table~\ref{supp_cityscapes} and Fig.~\ref{fig:supp_cityscapes} show that our proposed SWR greatly outperforms its baseline model and takes advantage of the strong baseline model, RobustNet, as a good starting point for test-time adaptation. In addition, our proposed SWR shows stable performance enhancement regardless of the learning rate in test-time adaptation, as shown in Table~\ref{supp_cityscapes_lr}.

\begin{figure}[!t]
\vspace*{0.2cm}
\centering
\includegraphics[width=\linewidth]{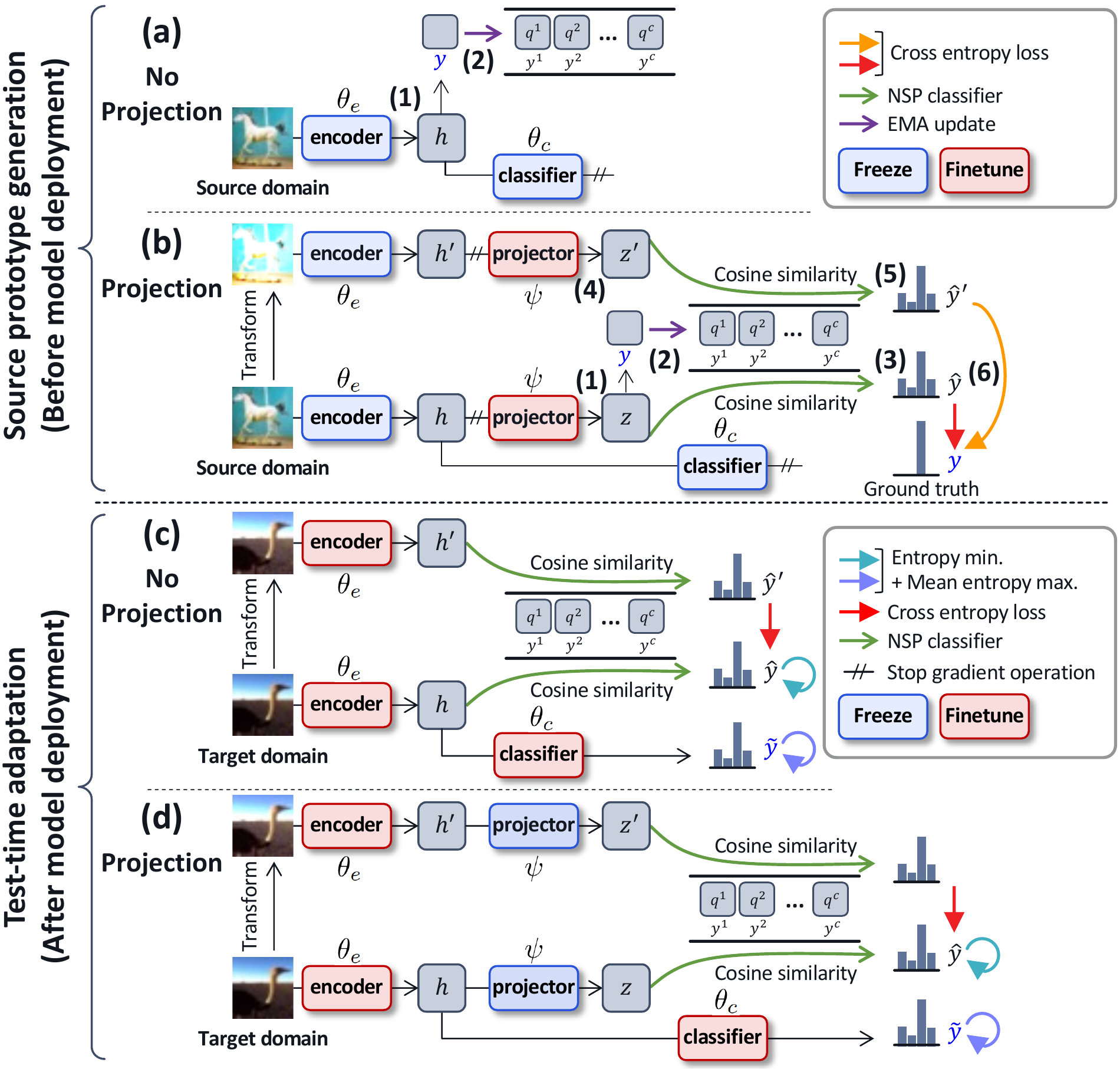}
\vspace*{-0.3cm}
\caption{Comparison of models with and without a projector. (a) and (c) describe the model without the projector, and (b) and (d) demonstrate the model with the projector. (a) The source prototypes are generated from the inferred feature representation $\boldsymbol{h}$ without going through the projector. (b) The source prototypes are generated from the inferred projection $\boldsymbol{z}$ through the projector. The auxiliary task loss is applied to the feature representation (c) or the embedding space behind the projector (d).}
\label{fig:supp_nsp}
\end{figure}

\begin{figure}[!t]
\vspace*{0.0cm}
\centering
\includegraphics[width=0.97\linewidth]{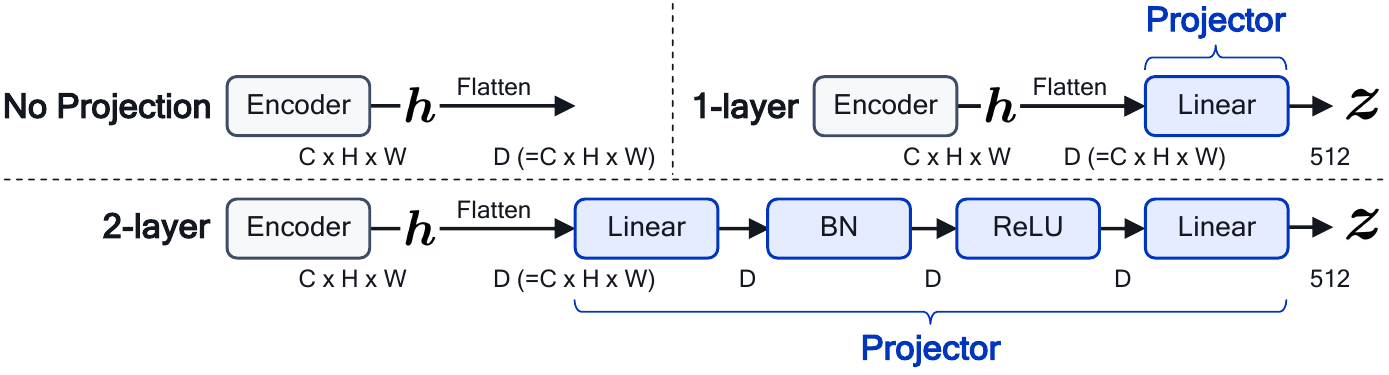}
\vspace*{-0.1cm}
\caption{Detailed architecture of the projector.}
\label{fig:supp_mlp}
\vspace*{-0.1cm}
\end{figure}

\section{Further Implementation Details}

\vspace*{-0.0cm}
\subsection{Detailed Projector Design}\label{supp_projector_design}
As described in Section~\textcolor{red}{3.3}, we attach and train a projector behind the encoder to map the feature representation $\boldsymbol{h}$ to the projection $\boldsymbol{z}$. The projector minimizes the misalignment between the source and target embeddings by enabling transformation-invariant mapping and bringing the projections belonging to the same class closer together in the new embedding space. However, in Section~\textcolor{red}{4.5}, we showed that applying auxiliary task loss directly to feature representation $\boldsymbol{h}$ without the projector may perform better on datasets with fewer classes (\textit{e.g.,} CIFAR-10-C). Fig.~\ref{fig:supp_nsp} shows the architectural differences between our proposed methods with and without the projector. If there is no projector, only steps (1) and (2) are repeated to obtain the prototype for each class by averaging over the feature representations $\boldsymbol{h}$ inferred across the source samples, as shown in Fig.~\ref{fig:supp_nsp}(a). Also, the auxiliary task loss is applied to the feature representation $\boldsymbol{h}$ that is the encoder's output without using the projector, as shown in Fig.~\ref{fig:supp_nsp}(c). Fig.~\ref{fig:supp_mlp} shows the detailed architecture of the projector.

\vspace*{-0.15cm}
\subsection{Experimental Setup for Domain Generalization Benchmarks}
As described in Section~\textcolor{red}{4.7}, our implementation uses DomainBed\footnote{https://github.com/facebookresearch/DomainBed}~\cite{gulrajani2020search} and T3A\footnote{https://github.com/matsuolab/T3A}~\cite{iwasawa2021test} framework to conduct the experiments on four domain generalization benchmarks such as PACS~\cite{li2017deeper}, OfficeHome~\cite{venkateswara2017deep}, VLCS~\cite{Fang_2013_ICCV}, and TerraIncognita~\cite{beery2018recognition}. Following T3A~\cite{iwasawa2021test}, we first train ResNet-50 backbone networks on each dataset by using ERM~\cite{vapnik1999overview} and CORAL~\cite{sun2016deep}. For this pre-training, we conduct a random search of five trials over the hyper-parameter distribution and repeat this procedure three times independently with a different random seed. Then, we apply our proposed method to each pre-trained model. Our method has two hyper-parameters: learning rate (LR) and projector. We set the search space for LR to $\text{LR}\in\{$5e-5, 1e-6$\}$ and provide two projector options: the model with or without the projector, as described in Section~\textcolor{red}{4.5} and~\ref{supp_projector_design}. Note that the hyper-parameter selection is completed before deployment to the test domain according to leave-one-domain-out cross-validation~\cite{gulrajani2020search}. We use a batch size of 200, and the other hyper-parameters described in Section~\textcolor{red}{4.2} are set the same as the experiments on CIFAR datasets.

%
%

\end{document}